\documentclass[10pt,twocolumn,letterpaper]{article}

\usepackage[dvipsnames]{xcolor}
\usepackage{iccv}
\usepackage{times}
\usepackage{epsfig}
\usepackage{graphicx}
\usepackage{amsmath}
\usepackage{subfigure}
\usepackage{amssymb}
\usepackage{booktabs}
\usepackage{comment}
\usepackage{caption}
\usepackage{caption}
\usepackage{float}
\usepackage{pifont}
\usepackage[ruled,vlined]{algorithm2e}
\usepackage{tabularx}
\usepackage[toc,page]{appendix}

\newcommand{\methodName}{\textsc{PixelPick}}
\newcommand{\cmark}{\ding{51}}%
\newcommand{\xmark}{\ding{55}}%

\DeclareMathOperator*{\argmin}{argmin}   
\DeclareMathOperator*{\argmax}{argmax}
\DeclareMathOperator*{\argmaxTwo}{argmax2}

\newcommand{\pascalShort}{\textsc{VOC12}}
\newcommand{\pascalLong}{\textsc{Pascal VOC 2012}}
\newcommand{\cv}{\textsc{CamVid}}
\newcommand{\cs}{\textsc{Cityscapes}}

\usepackage[pagebackref=true,breaklinks=true,letterpaper=true,colorlinks,bookmarks=false]{hyperref}

\iccvfinalcopy 

\def\iccvPaperID{3234} 

\ificcvfinal\fi

\begin{document}

\title{All you need are a few pixels: semantic segmentation with \methodName}

\author{Gyungin Shin \quad \quad \quad Weidi Xie \quad \quad \quad Samuel Albanie\\ [2pt]
Visual Geometry Group, Department of Engineering Science\\
University of Oxford, UK\\
{\tt\small \{gyungin, weidi, albanie\}@robots.ox.ac.uk}\\
{\small \url{https://www.robots.ox.ac.uk/~vgg/research/pixelpick}}
}

\maketitle
\ificcvfinal\thispagestyle{empty}\fi

\begin{abstract}
A central challenge for the task of semantic segmentation is 
the prohibitive cost of obtaining dense pixel-level annotations to supervise model training.
In this work, we show that 
in order to achieve a good level of segmentation performance,
all you need are a few well-chosen pixel labels.

We make the following contributions:
(i) We investigate the semantic segmentation setting in which labels are supplied only at sparse pixel locations, and show that deep neural networks can use a handful of such labels to good effect;
(ii) We demonstrate how to exploit this phenomena within an active learning framework, termed \methodName,
to radically reduce labelling cost, and propose an efficient ``mouse-free'' annotation strategy to implement our approach;
(iii) We conduct extensive experiments to study the influence of annotation diversity under a fixed budget,  model pretraining, model capacity and the sampling mechanism for picking pixels in this low annotation regime;
(iv) We provide comparisons to the existing state of the art in semantic segmentation with active learning,  
and demonstrate comparable performance with up to two orders of magnitude fewer pixel annotations on the \cv{}, \cs{} and \pascalLong{} benchmarks;
(v) Finally, we evaluate the efficiency of our annotation pipeline and its sensitivity to annotator error to demonstrate its practicality.
\end{abstract}
\section{Introduction} \label{sec:intro}
The coupling of deep neural networks and large-scale labelled datasets 
has yielded significant progress on a host of core machine perception tasks. 
A key challenge of training these models is their need for considerable quantities of annotation, 
which can be prohibitively expensive to collect for applications that require either specialised annotators (such as medical image diagnostics~\cite{Abramoff16,Gulshan16,Shih19,Wang2020}),
or fine-grained labels, such as for detection and segmentation~\cite{lin2014microsoft}.

Semantic segmentation, in particular, has proven valuable for decision making in a variety of applications such as digital pathology~\cite{tokunaga2019adaptive}, 
remote sensing~\cite{Wurm2019SemanticSO} and autonomous driving~\cite{yang2018denseaspp}. However, its requirement of per-pixel annotations raises significant scalability challenges---on average more than 1.5 hours of annotation and quality control
was required for each image in the \cs{} segmentation dataset~\cite{Cordts2016Cityscapes}.

\begin{figure}
    \centering
    \includegraphics[width=0.45\textwidth]{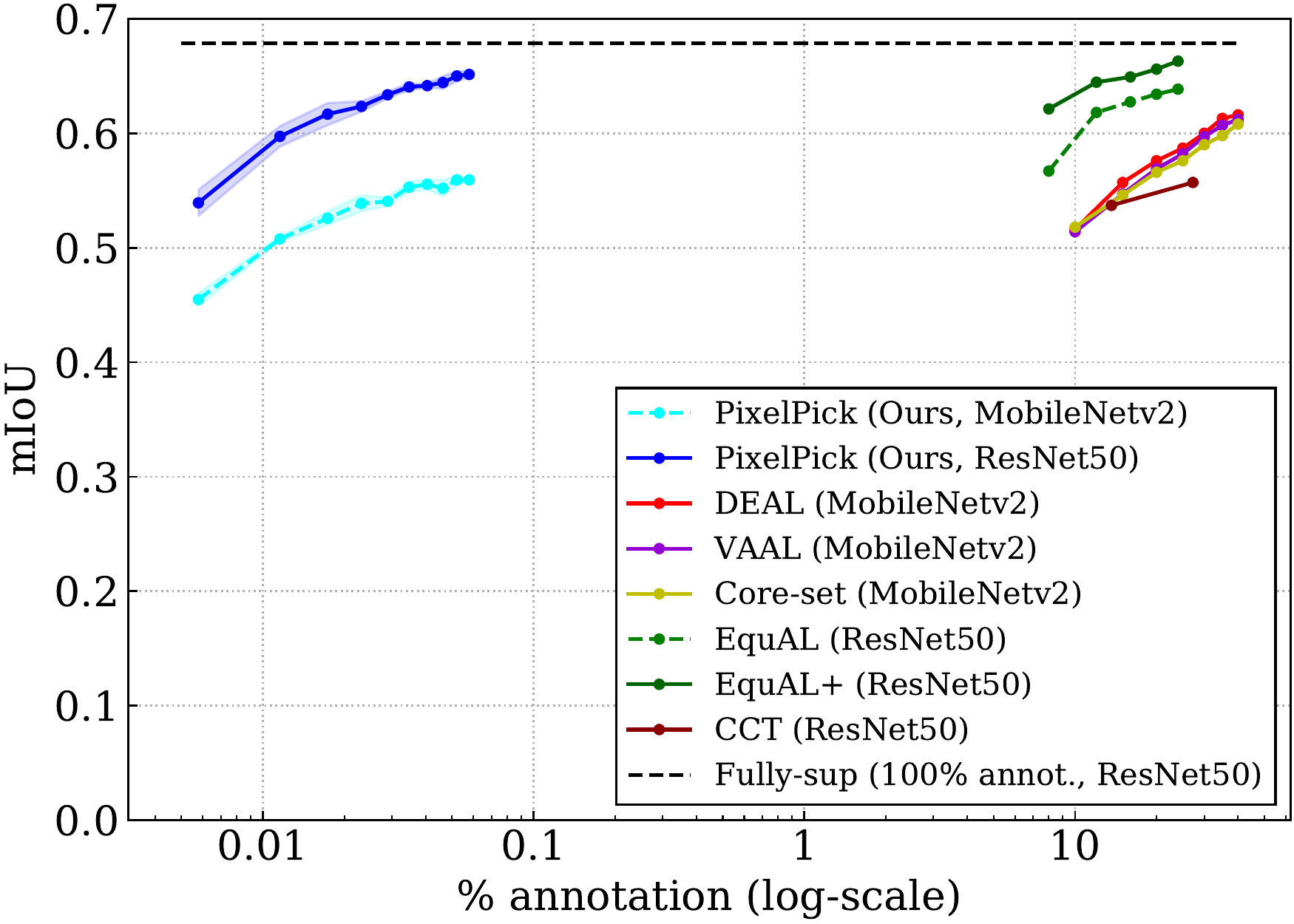}
    \vspace{-7pt}
    \caption{\textbf{All you need are a few pixels:} We show that deep neural networks can obtain remarkable performance with just a handful of labelled pixels per image whose spatial coordinates are proposed by the model, rather than the human annotator. We compare our approach, \methodName{}, with existing active learning and semi-supervised approaches on the \cv{} dataset~\cite{brostow2009semantic} 
    (see Sec.~\ref{sec:experiments} for further details).}
    \mbox{}\vspace{-0.6cm} \\
    \label{fig:teaser}
\end{figure}

The objective of this work is to propose a simple yet effective approach for training a good semantic segmentation model at minimal annotation cost.  
Our approach is motivated by three observations: 
(1) Within a given image, pixels exhibit significant spatial mutual information;
(2) Deep neural networks possess a strong inductive bias that renders them appropriate for modelling these spatial dependencies~\cite{ulyanov2018deep};
(3) Collecting mask, scribble or click annotations requires annotators to ``localise and classify'' using a mouse or trackpad. By contrast, assigning a class to a pixel proposal can be ``mouse-free'', requiring instead only a ``classify'' task without a localisation component (and which can be performed via a single key-press).  
The first two factors imply that densely labelling all pixels in images may be highly redundant, while the third suggests the possibility of designing an efficient sparse pixel labelling strategy.
Several questions then arise: 
\textit{how many sparse pixel labels are needed to achieve good performance?}
\textit{how should those pixel locations be selected?} and
\textit{how can the selected pixels be annotated efficiently?}

In this paper, we address these questions through the lens of \textit{active learning}~\cite{atlas1989training,Seung1992QueryBC}. 
In contrast to passive supervised learning (in which 
the model is tasked with learning a mapping from a fixed set of input-output pairs),
active learning considers a dynamic scenario in which a model can interactively request labels for the samples that it believes will be most useful for solving a given task. 
Our proposed \methodName{} framework adopts this paradigm,
learning a model for semantic segmentation by alternating between 
training on previously labelled pixels and requesting new pixel labels.

We make the following contributions:
(i) We study the problem setting in which labels are supplied at the level of sparse pixels and show that with only a small collection of such labels, modern deep neural networks can achieve good performance;
(ii) We show how this phenomenon can be exploited with an efficient and practical ``mouse-free'' annotation strategy as part of a proposed \methodName{} active learning framework;
(iii) We perform a series of experiments into factors that affect model performance in the low-annotation regime: annotation diversity, architectural choices and the design of the sampling mechanisms for selecting most useful pixels;
(vi) We compare with other state-of-the-art active learning 
approaches on standard segmentation benchmarks:
~\cv{}, \cs{} and \pascalLong{}, 
where we demonstrate comparable segmentation performance with significantly lower annotation budget~(Fig.~\ref{fig:teaser});
(v) Lastly, we assess \methodName{} from the perspective of practical deployment, assessing its annotation efficiency and robustness.
\section{Related work} \label{sec:related}
Our work is related to several themes of research that have sought to minimise labelling costs for semantic segmentation,
as discussed next. \\[-9pt]

\noindent \textbf{Weakly-supervised semantic segmentation}.
Many weak supervisory signals have been explored in the literature as a pragmatic compromise between fully supervised~\cite{long2015fully} and fully unsupervised approaches to semantic segmentation~\cite{ji2019invariant}. 
These cues include scribbles~\cite{lin2016scribblesup}, 
eye tracking~\cite{papadopoulos2014training}, 
object pointing~\cite{bearman2016s,qian2019weakly}, 
web-queried samples~\cite{jin2017webly}, 
bounding boxes~\cite{dai2015boxsup,khoreva2017simple,song2019box}, 
extreme clicks for objects~\cite{papadopoulos2017extreme,maninis2018deep} 
and image-level labels~\cite{zhou2016learning,wei2018revisiting,Fan_2018_ECCV}. Differently from these approaches, we gather labels at sparse pixel locations proposed by the model itself, rather than at locations selected by the annotator, and show that very few such annotations are needed for good performance.\\[-9pt]

\noindent \textbf{Interactive annotation}.
There is rich body of computer vision literature considering the related problem of accelerating \textit{interactive} annotation. 
The seminal work of~\cite{boykov2001interactive} demonstrated how to
exploit scribbles to indicate the foreground/background appearance model and leverage graph-cuts for segmentation~\cite{Boykov04}. This was later extended to the use of multiple scribbles on both object and background, 
applied to annotating objects in videos~\cite{Nagaraja15}.
\cite{rother2004grabcut} exploited 2D bounding boxes provided by the annotator
and performed pixel-wise foreground/background labelling using EM.
Recent work~\cite{PolyRNN}
tasks a model with sequentially producing the vertices of a polygon outlining an object, given an appropriate crop. As with the weakly-supervised signals described above, these methods are passive in the sense that the labelling process is driven by the human annotator, rather than the model.\\[-9pt]

\noindent \textbf{Semi-supervised semantic segmentation}.
Inspired by classical self-labelling approaches which aim to leverage unlabelled data to improve a classifier~\cite{scudder1965probability,yarowsky1995unsupervised}, a number of semi-supervised approaches have been developed to make use of pseudo-labelling algorithms~\cite{lee2013pseudo} for semantic segmentation in a low-annotation regime. Consistency-based pseudo-labelling methods have recently demonstrated promising results, highlighting the important role of aggressive data augmentations when only a small number of densely annotated images or regions are available~\cite{Ouali_2020_CVPR,french2019consistency}.

Our approach differs from theirs in several ways: 
(i) our model is trained from sparse pixel annotations, rather than a small number of densely labelled images, 
(ii) we employ active learning (samples are dynamically selected and queried for annotation by the model), 
which, as we show through experiments, brings additional improvements.  
We compare our approach quantitatively with theirs in Sec.~\ref{subsec:exp:sota}.\\[-9pt]

\noindent \textbf{Active learning for semantic segmentation}.
At its core, active learning is a set selection problem; the aim being to
determine the most informative subset of examples to acquire labels for,
given a labelling budget~\cite{atlas1989training,Seung1992QueryBC,Lewis1994ASA,Freytag2014SelectingIE,Yoo2019LearningLF}.
In this case the maximally informative labelled-pixel
subset is the one which yields the lowest generalization error when used to train
a supervised semantic segmentation model. Prior work targetting segmentation has investigated strategies to select superpixels that induce the maximum label change for a CRF on the training set by using weak (image-level category) supervision~\cite{vezhnevets2012active},
incorporate geometric constraints~\cite{konyushkova2015introducing,mosinska2016active} 
and propagate foreground masks to large-scale image collections~\cite{jain2016active}.
For foreground segmentation of medical imagery, 
FCNs~\cite{long2015fully} have been coupled with bootsrapping~\cite{yang2017suggestive}, 
and U-Nets~\cite{ronneberger2015u} with dropout-based Monte Carlo estimates of uncertainty~\cite{gorriz2017cost} to drive label acquisition via uncertainty sampling.
The strategy of learning an estimator for difficult regions~\cite{Yoo2019LearningLF} has proven effective as a basis for selecting which images should be densely labelled for semantic segmentation~\cite{Xie2020DEALDA}. 

More closely related to our work, 
prior studies have considered \textit{region-based} sampling strategies for semantic segmentation, employing reinforcement learning~\cite{casanova2020reinforced}, 
equivariance constraints~\cite{golestaneh2020importance} 
and learned estimators of labelling cost~\cite{mackowiak2018cereals}. 
In contrast to these lines of research, 
our work aims to introduce a more efficient paradigm of active learning for segmentation, 
which is to train models with only sparse pixel annotations~(removing the localisation component of the annotation task).
We compare our approach with theirs in Sec.~\ref{sec:experiments}. \\[-9pt]
\vspace{-.1cm}
\section{Method} \label{sec:method}
\begin{figure*}[t]
    \centering
    \includegraphics[clip,trim={0cm 8cm 2cm 0cm},width=\textwidth]{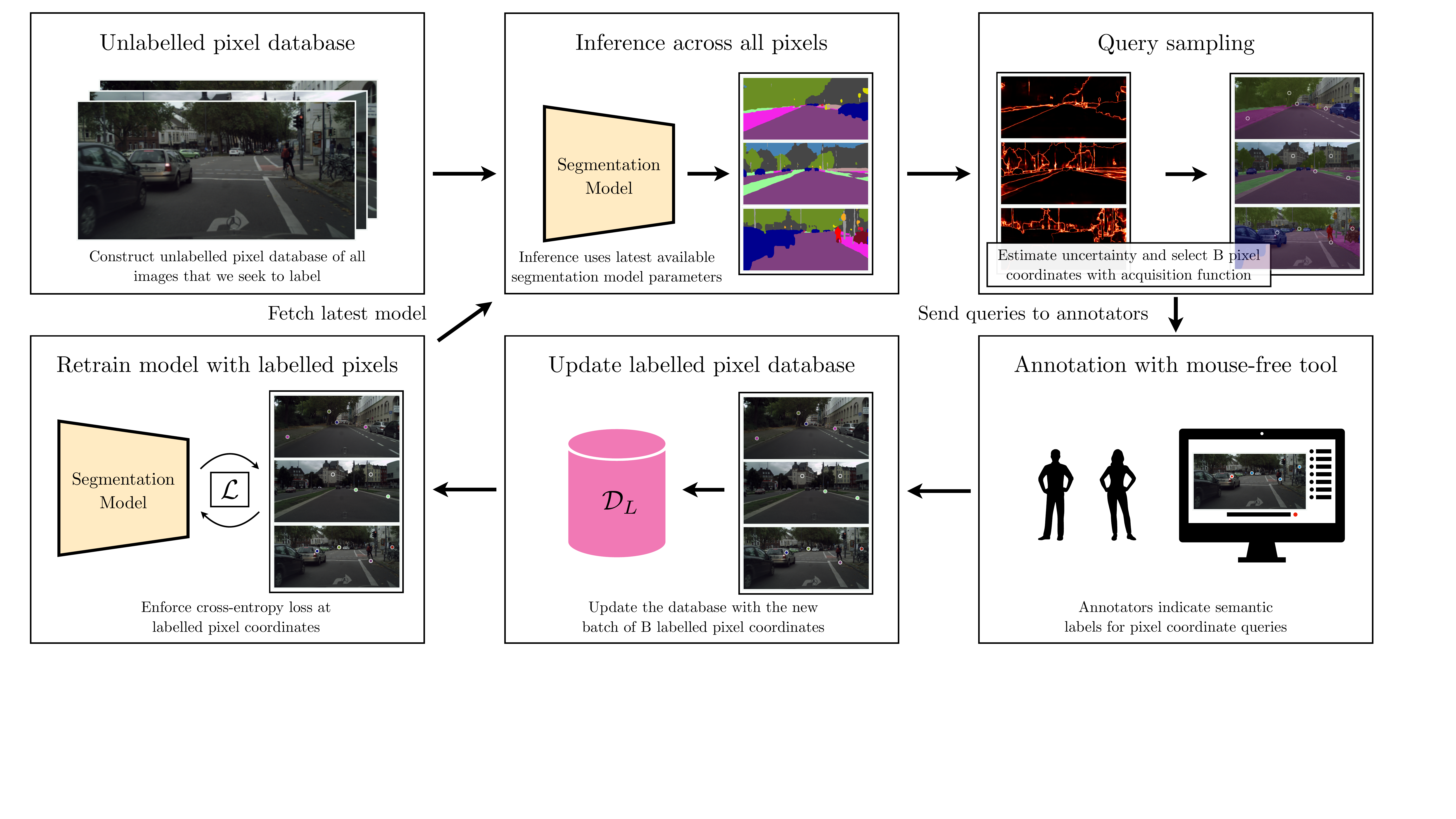}
    \vspace{-.4cm}
    \caption{\textbf{Overview of the \methodName{} active learning framework.} 
    Given a database of unlabelled pixels of interest (top-left) each image is fed to a segmentation model to produce pixel-wise class probabilities (top-middle), which are in turn passed to an acquisition function to estimate per-pixel uncertainties and select a batch of $B$ pixels to be labelled (top-right). The queries are sent to annotators (bottom-right), and the resulting labels are added to the \textit{labelled pixel database}, $\mathcal{D}_L$ (bottom-middle). Finally, the segmentation model is retrained on the expanded database (bottom-left), before the cycle repeats. To bootstrap the process and train the initial segmentation model, we randomly sample $B$ pixels and send them to be annotated. See text in Sec.~\ref{sec:method} for further details.}
    \label{fig:framework}
    \vspace{-.5cm}
\end{figure*}

In this section, 
we describe the problem formulation and introduce our framework for pixel-level active learning semantic segmentation in Sec.~\ref{subsec:framework}.
We then detail our mouse-free annotation tool to efficiently implement the framework in Sec.~\ref{subsec:annotationTool}.

\subsection{\methodName{} framework} \label{subsec:framework}
We seek to train a model for semantic segmentation with \textit{pool-based active learning}~\cite{settles2009active},
in which we alternate between training a model on available annotation and requesting new labels for unlabelled samples from an oracle (see Fig.~\ref{fig:framework}). 

More formally, let $\mathcal{X} \subset \mathbb{R}^{H \times W \times 3}$ denote the space of colour images and
let $\Phi(\cdot; \Theta): \mathcal{X} \rightarrow \mathcal{Y}^{H \times W}$ represent a ConvNet with parameters $\Theta$ that maps a given image to a grid of elements in a $C$-class semantic label space (here $\mathcal{Y}$ denotes the ($C-1$)-simplex, i.e. $\mathcal{Y} = \{ (p_1, \dots, p_C) \in [0, 1]^C : \sum_{p=1}^C p_i = 1, p_i \geq 0\}$).
We assume access to an initial unlabelled pool of $N$ images, $\mathcal{D}_U$, 
indexed by the $H \times W \times N$ pixel coordinate lattice, 
$\Omega$, and an annotation database, $\mathcal{D}_L^0$, initialized to be empty.

At the $k^{\mathrm{th}}$ round of learning, a batch of $B \in \mathbb{N}$ pixel coordinates, $\omega_k \subset \Omega$, are sampled by an acquisition function, $\mathcal{A}$, using the predictions of the model trained in the previous round, $\Phi(\cdot; \Theta_{k-1})$, on the unlabelled pool $\mathcal{D}_U$, i.e. $\mathcal{A}(\mathcal{D}_U, \Phi(\cdot; \Theta_{k-1})) = \omega_k \subset \Omega$. The sampled pixel coordinates $\omega_k$ are then sent to an oracle for labelling to  produce a corresponding set of one-hot labels $\{ y_u \in \mathcal{Y} : u \in \omega_k \}$ that are added to the latest version of the annotation database, $\mathcal{D}_L^{k-1}$. Finally, the model is retrained on this expanded database, $\mathcal{D}_L^k = \cup_{i=1}^k \{(u, y_u) : u \in \omega_i \} $ (comprising all annotations gathered so far), to produce a new model, $\Phi(\cdot; \Theta_k)$, and the process is repeated. We term this framework \methodName{} due to its emphasis on selecting appropriate pixels for annotation. The two components of the framework, namely \textit{retraining the segmentation model} and \textit{sampling new pixel coordinates}, are discussed next.\\[-9pt]

\par{\noindent \textbf{Retraining the segmentation model.}}
At round $k$ of the active learning algorithm, we solve for parameters $\Theta_k$ by minimising a cross-entropy loss at each labelled pixel coordinate present in the current annotation database $\mathcal{D}_L^k$:
\begin{align}
    \Theta_k &= \argmin_{\Theta} \mathcal{L}(\Theta, \mathcal{D}_L^k) \quad \text{      where} \\
    \mathcal{L}(\Theta, \mathcal{D}_L^k) &= -\frac{1}{|\mathcal{D}_L^k|} \sum_{(u, y_u) \in \mathcal{D}_L^k}^n \sum_{c=1}^C y_u(c) \cdot \log(\hat{y}_u(c)).
\end{align}
In the expression above, $y_u(c)$ and  and $\hat{y}_u(c)$ denote the $c^{\mathrm{th}}$ channel of the oracle-provided label and corresponding model prediction at pixel coordinate $u$, respectively. \\[-9pt]

\par{\noindent \textbf{Sampling new pixel coordinates for labelling.}}
The objective of the acquisition function, $\mathcal{A}$, is to sample the $B$ pixel locations at round $k$ that maximise improvement in segmentation performance for the current model $\Phi(\cdot; \Theta_{k-1})$.  Functionally, it acts by examining the predictions of $\Phi(\cdot; \Theta_{k-1})$ across all candidate pixel coordinates among the unlabelled pool $\mathcal{D}_u$ and sampling $B$ such coordinates according to a specified criterion. \\[-9pt] 

\par{\noindent \textbf{Discussion. }} 
The distinction between sampling contiguous spatial patches for annotation~(\textit{e.g.}~grids of 128x128 pixels or larger as considered in prior work~\cite{mackowiak2018cereals,golestaneh2020importance,casanova2020reinforced}), 
and sampling \textit{individual pixel coordinates}, 
as proposed within the \methodName{} framework, is a subtle but important one. 
It has two key benefits. 
The first, as noted in Sec.~\ref{sec:intro}, 
is that it allows us to leverage the powerful inductive biases provided by deep neural network architectures 
that render them well suited to modelling spatial dependencies in natural images~\cite{ulyanov2018deep}. 
The second is a practical one: by providing annotators with pixel coordinate proposals, 
the labelling process is transformed from a ``localise and classify'' task (required when segmenting semantic regions and typically performed with a mouse or trackpad), 
into simply a ``classify'' task in which a class label is assigned to a coordinate proposal, 
and which can often be performed with a single key-press. 
We validate both claims through experiments in Sec.~\ref{sec:experiments}, 
where we show that (i) deep neural networks achieve strong segmentation performance at extremely sparse annotation levels, 
(ii) ``mouse-free'' annotation can be performed very efficiently. \\[-9pt]

\par{\noindent \textbf{Acquisition functions. }} 
The design of the specific criteria employed by the acquisition function has been the subject of considerable attention in the active learning literature~(see~\cite{settles2009active} and~\cite{Ren2020ASO} for surveys of classical and recent approaches, respectively).  
Since the focus of our work is not the design of another criterion, 
but rather on the effectiveness of individual pixels as the base unit for annotation, 
we consider several existing approaches based on the framework of \textit{uncertainty sampling}~\cite{Lewis1994ASA} that have been noted as effective in the literature, discussed next.

The \textit{Least Confidence} acquisition strategy~\cite{lewis1994heterogeneous,Culotta2005ReducingLE}
draws, at each iteration, the pixel coordinate for which the model has \textit{least confidence} in its \textit{most likely} class label:
\begin{align}
    u_{LC}^* = \argmin_{u \in \Omega} \argmax_{c\in \{1, \dots, C\}} \hat{y}_u(c).
\end{align}

The \textit{Margin Sampling} strategy~\cite{Scheffer2001ActiveHM} looks for samples that exhibit the smallest difference (i.e. lowest ``margin'') between the first and second most probable labels:  
\begin{align}
	u_{MS}^* = \argmin_{u \in \Omega} \Big( \argmax_{c_1\in \{1, \dots, C\}} \hat{y}_u(c_1) - \argmaxTwo_{c_2\in \{1, \dots, C\}} \hat{y}_u(c_2)\Big),
\end{align} 
where the notation $\argmax2$ denotes the argument with the second largest value. 
Intuitively, pixel coordinates with small margins are ambiguous for the classifier, while those with large margins represent samples for which the classifier has greater confidence in its correctness.

Finally, the \textit{Entropy Sampling} strategy aims to select the pixel coordinate with the greatest conditional entropy~\cite{Shannon1948AMT} under the current model:
\begin{align}
    u_{ENT}^* = \argmax_{u \in \Omega} -\sum_{c=1}^C \hat{y}_u(c) \log \hat{y}_u(c).
\end{align}

As noted in prior work~\cite{Beluch2018ThePO,Yoo2019LearningLF}, 
these strategies can suffer from a lack of diversity if applied naively, 
but can be readily adapted to minimise this effect by first sub-sampling the unlabelled pool and then employing the acquisition function to choose only from this restricted subset. A variation of this diversity heuristic worked well on our task:
We first rank all pixels using the acquisition function,  
then uniformly sampling $B/N$ pixel coordinates from the top $M\%$ ranked locations in each image, where $M$ is a hyperparameter
and $N$ denotes the number of images we distribute our budget $B$ amongst.
We note that while more sophisticated strategies (e.g.~\cite{Sener2018ActiveLF}) 
could also be considered within our framework, 
a simple \textit{Margin Sampling} strategy coupled with the modification described above proved effective (shown through experiments in Sec.~\ref{sec:experiments}), and thus we adopt it in this work.\\[-9pt]

\noindent \textbf{Sampling batches.} 
The number of pixel coordinates sampled in each round, $B$, is set as a hyperparameter. 
A larger value of $B$ corresponds to fewer rounds of annotation (and therefore a potentially faster deployment cycle), 
at some cost in performance. 
A detailed study of the effects of $B$ is provided in the suppl. mat.

\subsection{\methodName{} Annotation tool} \label{subsec:annotationTool}
\begin{figure}
    \centering
    \includegraphics[width=0.46\textwidth]{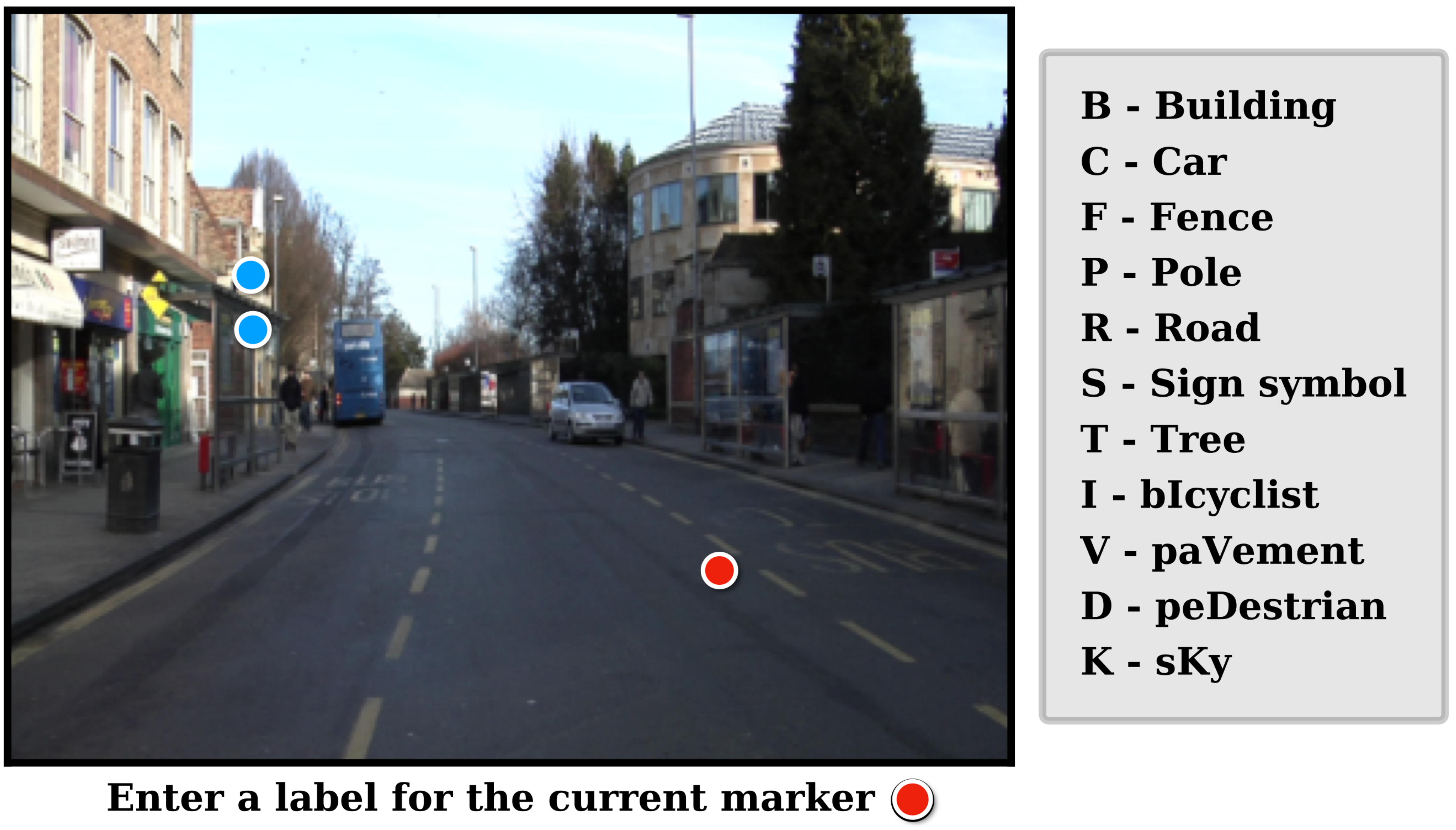}
    \vspace{-5pt}
    \caption{\textbf{\methodName{} mouse-free annotation tool.} The annotator classifies the highlighted point (in red) by pressing the keyboard character of the corresponding class for the dataset. The tool then highlights the next pixel proposal and the process repeats. Note that the task involves classification, but not localisation.
    }
    \label{fig:mf-tool}
\vspace{-.5cm}
\end{figure}

To demonstrate the practical utility of the \methodName{} framework, 
we created an annotation tool to support the labelling process (Fig.~\ref{fig:mf-tool}). 
The tool is simple: 
for each image, the annotator is presented with a few pixels that were selected by the \methodName{} acquisition function (described in Sec.~\ref{subsec:framework}). 
They are also shown a mapping from keyboard keys to semantic labels (Fig.~\ref{fig:mf-tool}, right hand side). 
The tool iterates over the pixel locations, 
highlighting the current pixel in red and the annotator simply presses the appropriate key to classify it. 
The tool then moves on to the next pixel proposal, 
and the procedure repeats until all proposals in the image are exhausted, when a new image is shown. 

We note that an important difference between this annotation technique and those considered in prior work (e.g. scribbles~\cite{lin2016scribblesup}, object pointing~\cite{bearman2016s,qian2019weakly}, extreme clicks~\cite{papadopoulos2017extreme,maninis2018deep} etc.) 
is that it is ``mouse-free''---requiring only key presses from the user 
---but avoids the complexity of specialised approaches such as eye tracking~\cite{papadopoulos2014training}.  
In Sec.~\ref{sec:experiments}, we conduct experiments to validate the efficiency of the proposed annotation tool.

\section{Experiments} \label{sec:experiments}
In this section, we first describe the datasets used in our experiments in Sec.~\ref{subsec:exp:datasets} before providing implementation details in Sec.~\ref{subsec:exp:imp}. In Sec.~\ref{subsec:exp:ablations}, we conduct extensive ablation studies, and we then compare with existing state-of-the-art approaches in Sec.~\ref{subsec:exp:sota}.
Finally, in Sec.~\ref{subsec:exp:practical},
we demonstrate the practical feasibility and robustness of \methodName{} by reporting annotation times
and investigating its sensitivity to annotator errors.

\subsection{Datasets} \label{subsec:exp:datasets}

\par{\noindent \textbf{\cv{}}~\cite{brostow2009semantic}}
is an urban scene segmentation dataset composed of $11$ categories and 
containing $367$, $101$, and $233$ images of $360\times480$ resolution for training, validation, and testing, respectively. 
\\[-9pt]

\par{\noindent \textbf{\cs{}}~\cite{Cordts2016Cityscapes}}
is a dataset collected for the purpose of autonomous driving 
consisting of $2975$ training, $500$ validation and $1525$ test high-resolution images ($1024\times2048$) with 19 classes. 
During training, we resize the images to $256\times512$ pixels to make the training time manageable, 
and perform inference on images of $512\times1024$ pixels. \\[-9pt]

\par{\noindent \textbf{\pascalLong{}}~\cite{Mark2015VOC}} (abbreviated to \pascalShort{})
contains $1464$, $1449$, and $1456$ images for training, validation and testing respectively. Each pixel is labelled as one of the 20 semantic categories or background. Since images in this dataset have different sizes, during training we resize the larger image dimension to 400 and randomly crop a $320\times320$ patch as input, and use the original size for inference, following \cite{Ouali_2020_CVPR}.

\begin{figure*}[!htb]
  \centering
  \hspace{-20pt}
  \subfigure[Effect of depth (\cv)]
  { 
    \label{fig:exp:depth_camvid}
  \includegraphics[width=0.248\textwidth,height=0.181\textwidth]{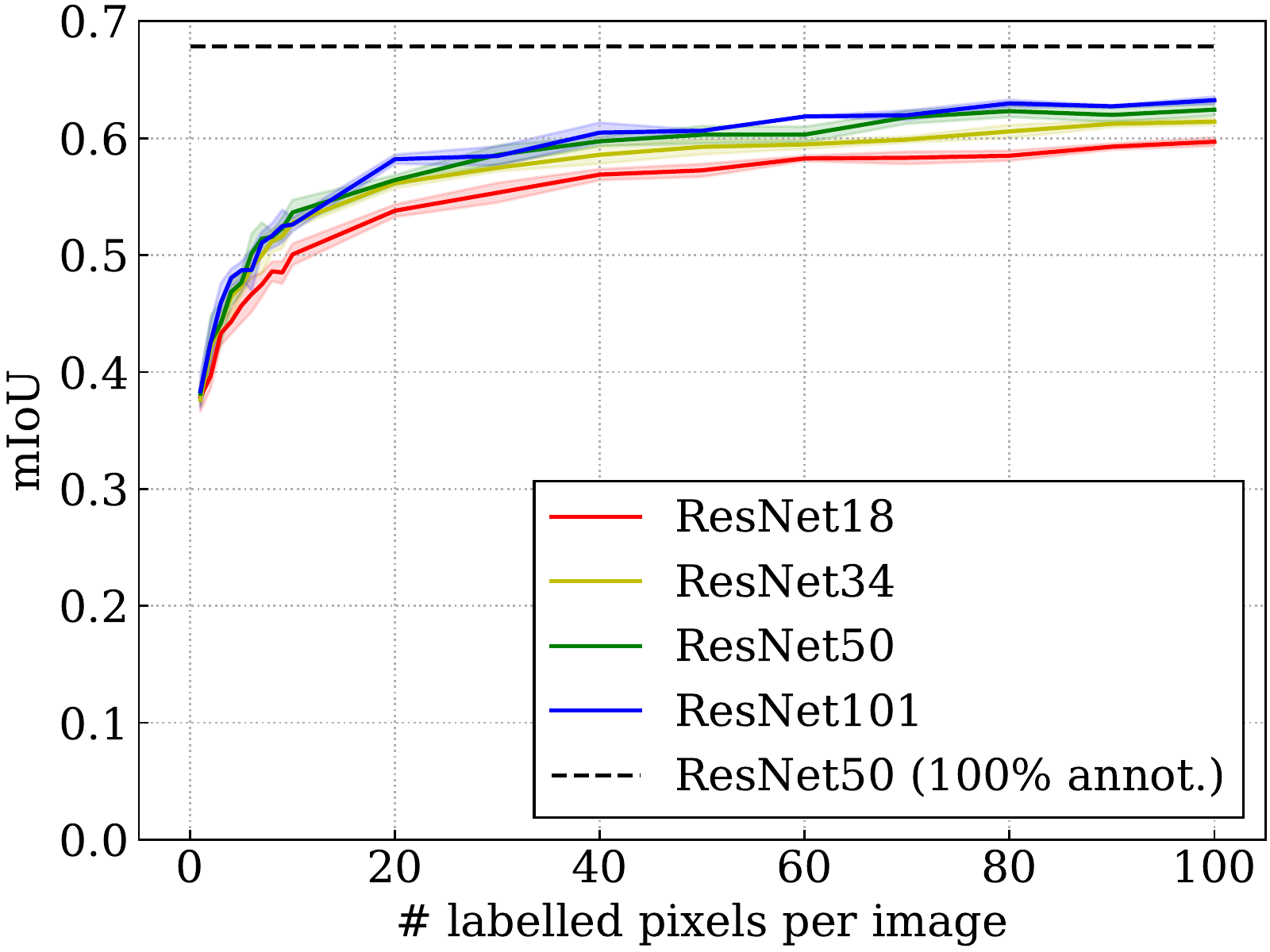}%
  }%
  \subfigure[Effect of depth (\pascalShort)]
  { 
    \label{fig:exp:depth_pascal}
  \includegraphics[width=0.248\textwidth,height=0.175\textwidth]{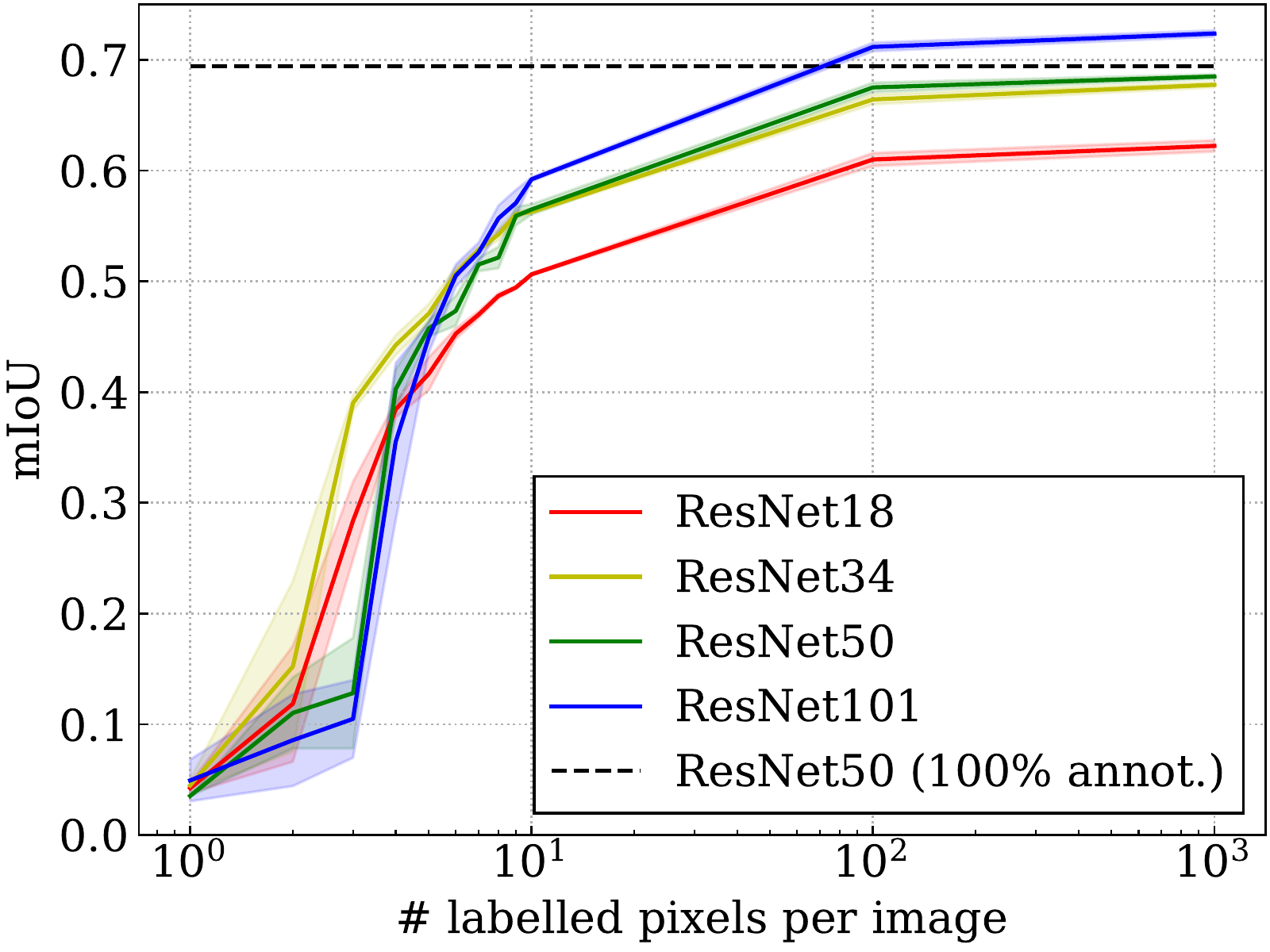}%
  }%
  \hspace{1pt}
  \subfigure[Effect of pretraining (\cv)]
  {  
  \label{fig:exp:self_sup_camvid}
  \includegraphics[width=0.248\textwidth,height=0.174\textwidth]{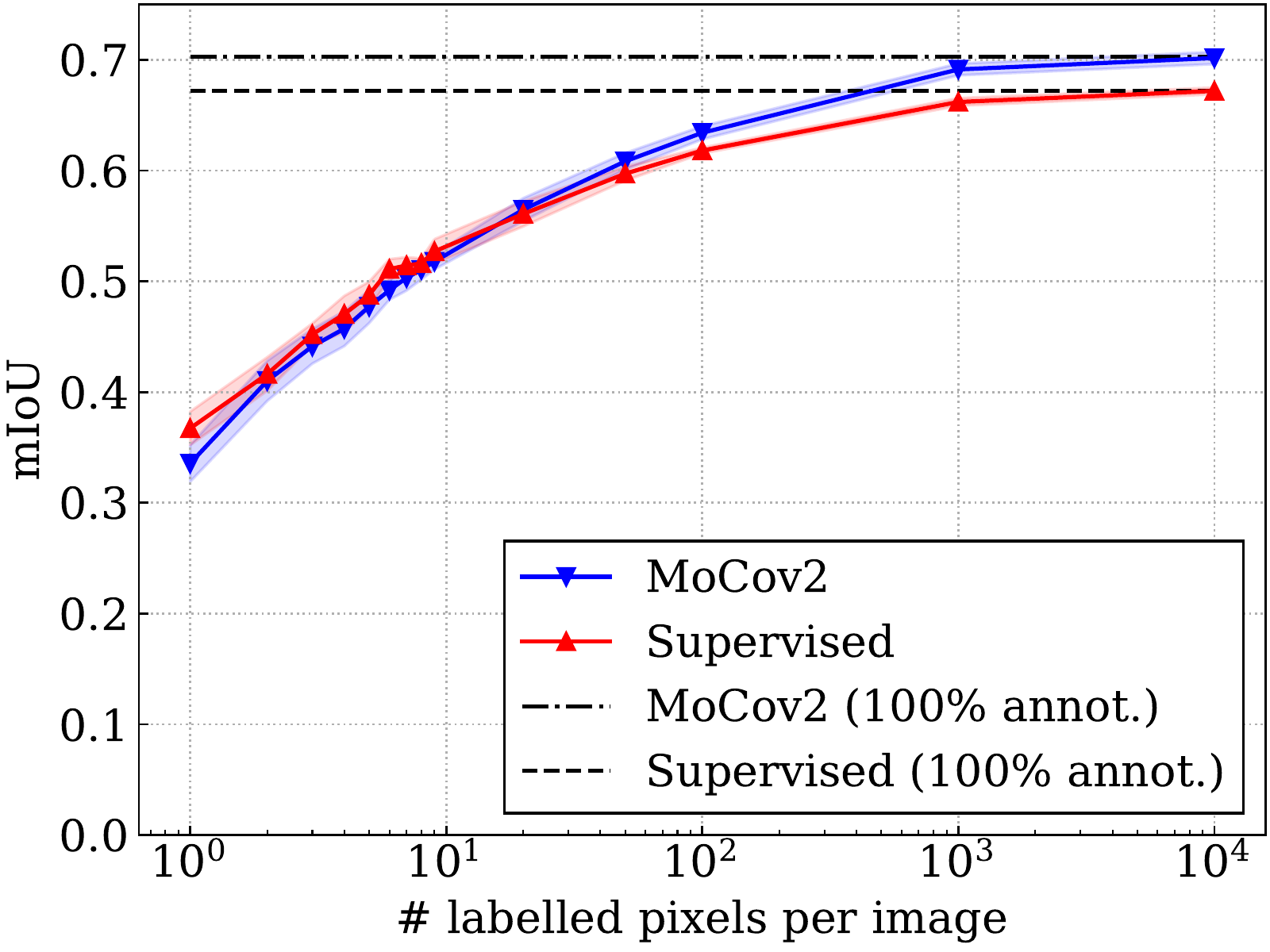}
  }
  \hspace{-5pt}
  \subfigure[Effect of pretraining (\pascalShort)]
  { 
    \label{fig:exp:self_sup_pascal}
  \includegraphics[width=0.248\textwidth,height=0.173\textwidth]{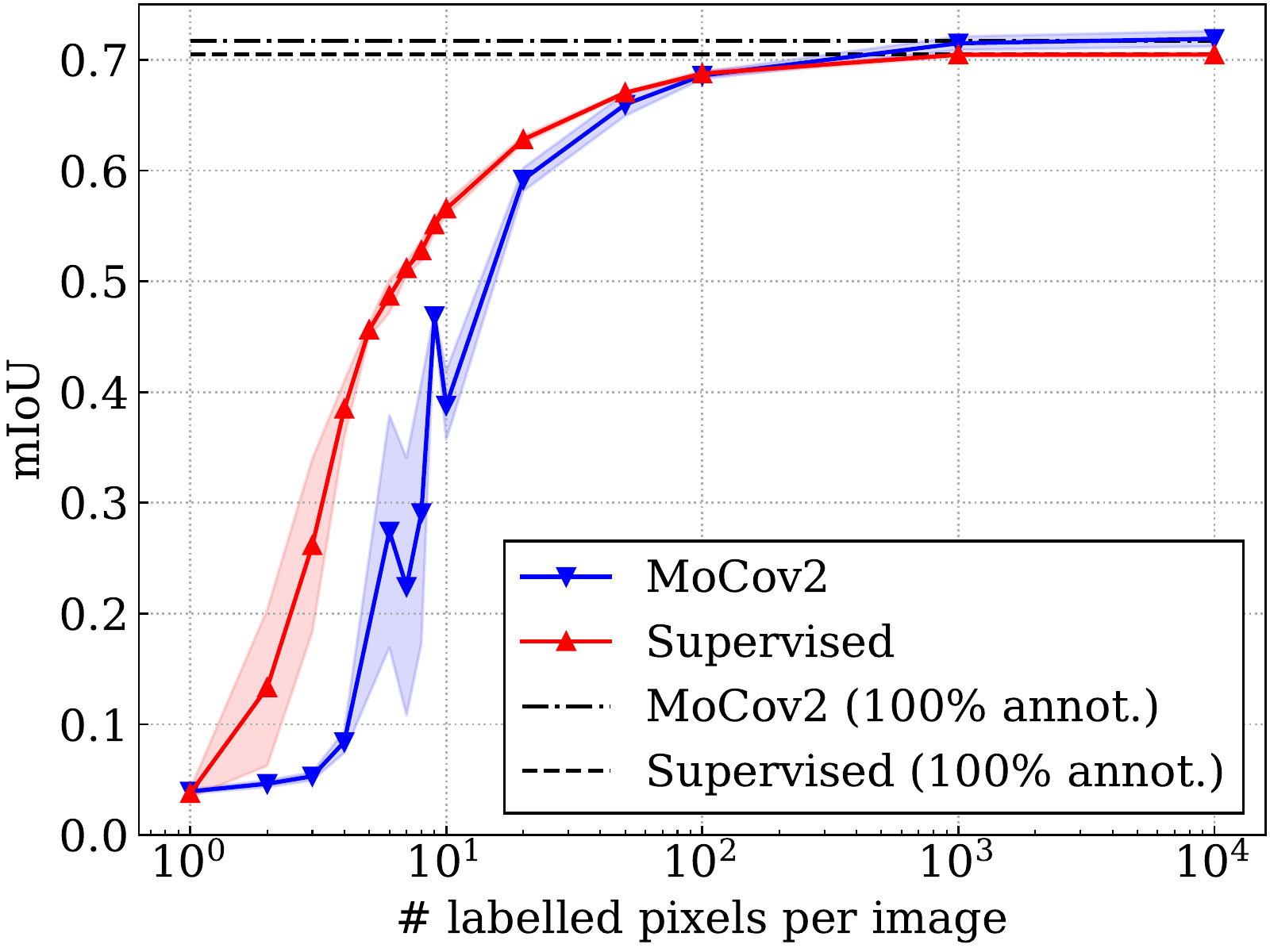}%
  }
  \vspace{-.2cm}
  \caption{\textbf{Ablation studies.} In (a) and (b), we investigate the effect of segmentation encoder \textit{depth} on \cv{} and \pascalShort, respectively. We observe that greater depth consistently helps performance above a threshold of 10 pixel labels per image.
  In (c) and (d), we compare fully-supervised ImageNet classification pretraining with self-supervised ImageNet (MoCov2~\cite{chen2020improved}) pretraining for the encoder on \cv{} and \pascalShort{}, respectively, where we see that for lower numbers of pixel labels per image that fully-supervised pretraining is a better choice, but the situation reverses as more annotations become available.}
  \vspace{-0.3cm}
\end{figure*}

\subsection{Implementation details} \label{subsec:exp:imp}
\noindent \textbf{Network architectures.}
We adopt two architectures for our experiments. For a lightweight model, 
we use DeepLabv3+~\cite{Chen_2018_ECCV} with MobileNetv2~\cite{Sandler2018MobileNetV2IR} as the backbone, following~\cite{Siddiqui2020ViewAL, Xie2020DEALDA}.
We also consider a heavier model for ablations and for comparison with the existing state-of-the-art approaches: a Feature Pyramid Network (FPN)~\cite{Lin2017FeaturePN} with a dilated version of ResNet50~\cite{He_2016_CVPR}, that replaces the stride of the last two residual blocks with atrous convolutions following~\cite{Ahn_2018_CVPR, huang2018dsrg, Ouali_2020_CVPR, Zhao_2017_CVPR}.  \\[-9pt]

\noindent \textbf{Training details.}
During each round of active learning,
we enforce the cross-entropy loss only on the labelled pixels (i.e.~those in $\mathcal{D}_L^k$ for round $k$), as described in Sec.~\ref{subsec:framework}.  Unless otherwise stated, $M$, the hyperparameter defining the \% of top ranked pixel coordinates used as a basis for uniform sampling is set to 5, while $B$, the pixel labelling budget per round is set to $10N$ for \cv{} and \cs{} and $5N$ for \pascalShort, where $N$ is the number of images in the dataset. At the beginning of each round, we reinitialise the model and train from scratch with the updated labelled pixels. For optimisation, we use Adam~\cite{2014arXivKingma} with a learning rate of $5\times 10^{-4}$ for the \cv{} and \cs{} datasets,
and SGD with momentum 0.9 and a learning rate of $10^{-2}$ for the \pascalLong{} dataset. 
For \cv,  we train for 50 epochs and decay the learning rate at 20 and 40 epochs by a factor of 10. 
On \cs{} and \pascalLong{}, we use Poly learning rate schedule as in~\cite{Ouali_2020_CVPR, Xie2020DEALDA, Chen_2018_ECCV, 2015LiuarXiv}. 
For data augmentation, we largely follow \cite{Ouali_2020_CVPR}, 
and use random scaling between [0.5, 2.0] and random horizontal flipping. 
In addition, we apply photometric transformations such as colour jittering, 
random grayscaling and Gaussian blurring. \\[-9pt]

\par{\noindent \textbf{Evaluation metrics.}}
Following standard practice~\cite{golestaneh2020importance, Xie2020DEALDA, Ouali_2020_CVPR, mackowiak2018cereals}, 
we compute mean intersection over union (mIoU), report our results on the test set for \cv{}, 
and on the validation set for \cs{} and \pascalShort{} datasets.
To provide a measure of variance in our low data regime, 
we report the average of 3 different runs (i.e., different seeds) on \pascalLong{} and 5 runs on \cv{} and \cs{} for all experiments. We plot their standard deviations as shaded regions ($\pm1$ std. dev.). \\[-9pt]

\subsection{Ablation studies} \label{subsec:exp:ablations}
In this section, 
we explore the effect of four factors that affects 
the performance in the \methodName{} framework,
with a particular focus on the small $B$ setting.
\textit{annotation diversity} (with the goal of finding the most effective way to spend an annotation budget);
\textit{encoder depth} (varying the capacity of the encoder);
\textit{encoder initialisation}~(self-supervised vs supervised pretraining);
and \textit{acquisition function} (determining the best way to select pixels). 
Note that, 
while investigating the first three factors, 
all pixels are selected via simple uniform random sampling, 
with the goal of validating the effectiveness of inductive bias in modern ConvNets.
We simulate the active learning process, 
following standard practice~\cite{golestaneh2020importance, Xie2020DEALDA, casanova2020reinforced},
\textit{i.e.}~to label the queried pixels, 
we simply reveal labels by querying the ground truth annotations at their spatial coordinates. \\[-9pt]

\begin{figure*}[!htb]
  \centering
  \subfigure[Effect of annotation diversity]{
  \label{fig:exp:data_diveristy}
  \hspace{-20pt}
  \includegraphics[width=0.33\textwidth,height=0.191\textwidth]{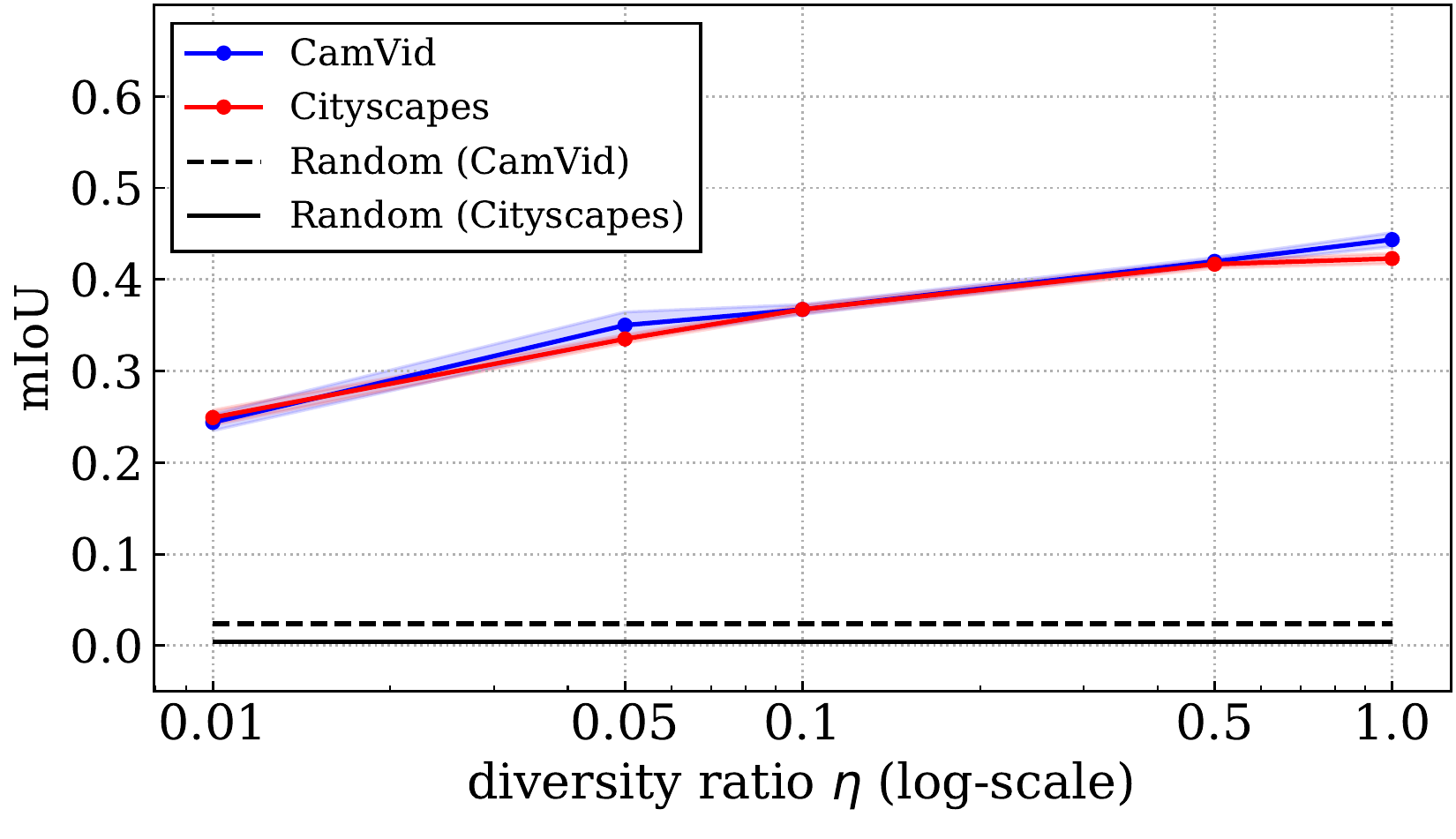}
  }
  \hspace{-5pt}
  \subfigure[Acquisition functions (\cv)]{
  \label{fig:exp:uncertainty}
  \includegraphics[width=0.33\textwidth,height=0.19\textwidth]{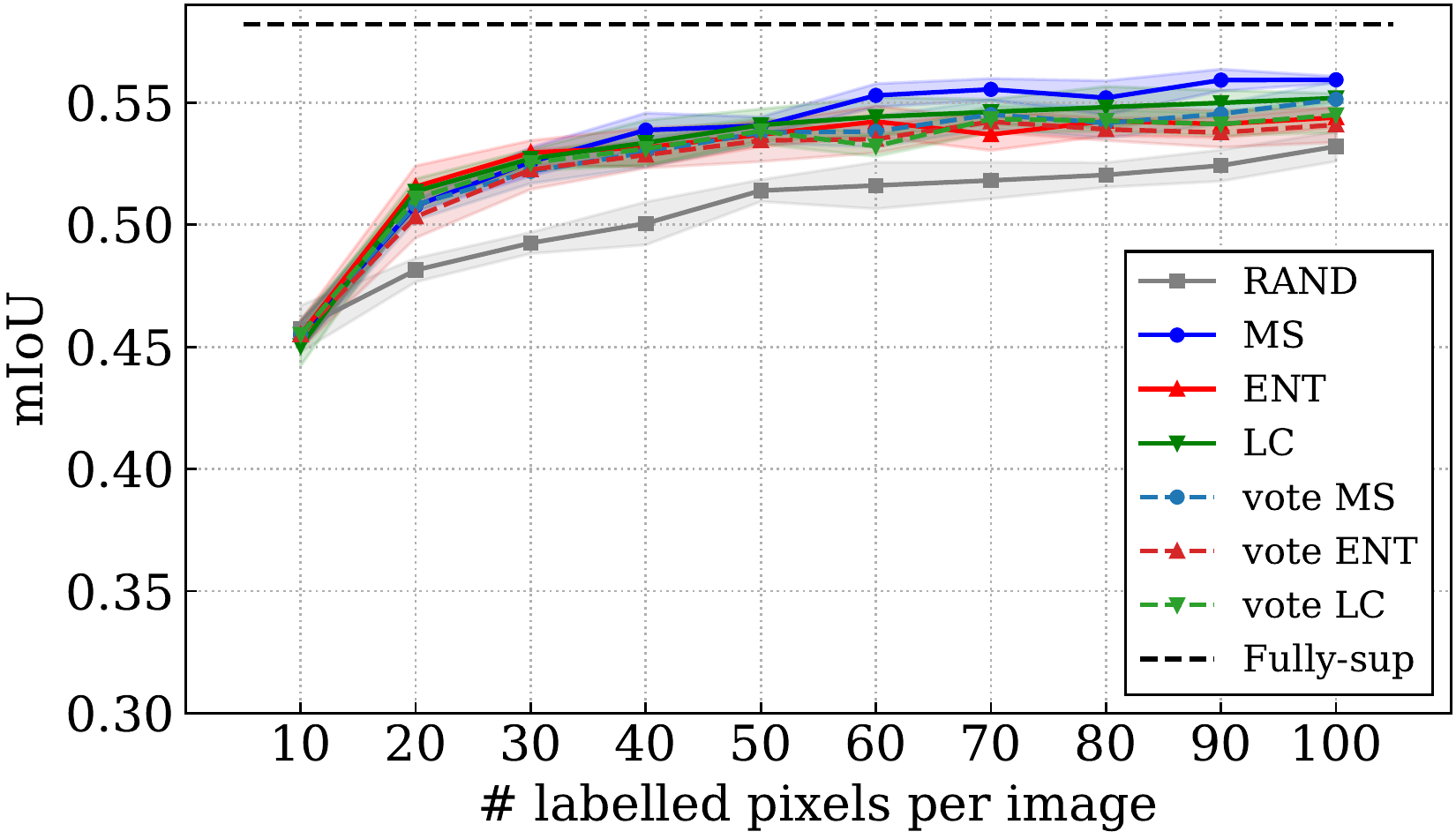}%
  }%
  \hspace{-2pt}
  \subfigure[Sensitivity to annotator error (\pascalShort)]{
  \label{fig:exp:errsim}
  \includegraphics[width=0.33\textwidth,height=0.188\textwidth]{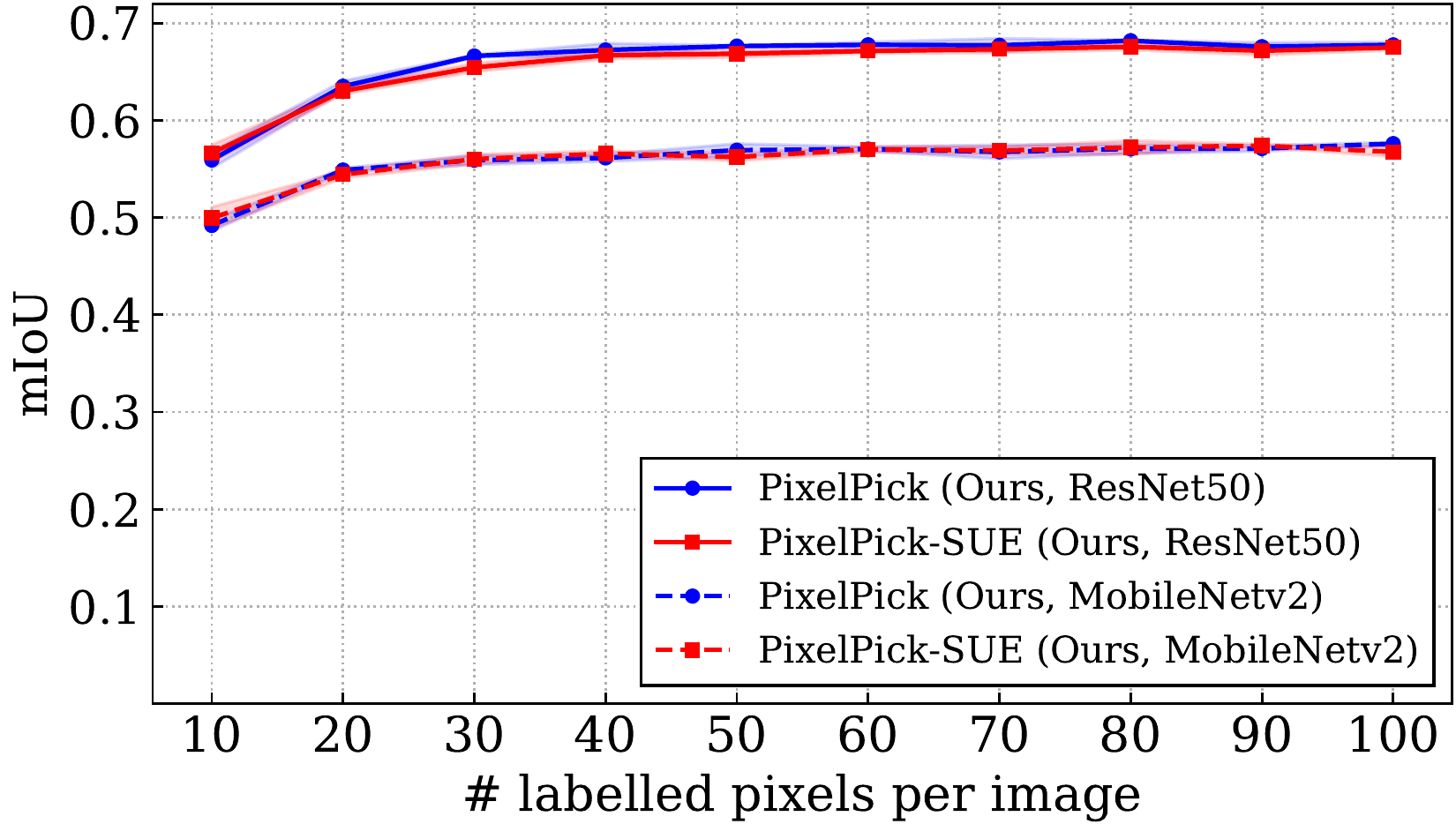}
  }
  \vspace{-.2cm}
  \caption{\textbf{Ablation studies.} In (a), we observe that sparsely annotating a larger number of images (higher $\eta$ value) outperforms denser labelling of fewer images, with consistent trends on the \cv{} and \cs{} datasets. In (b), we compare acquisition functions on \cv{} and find that \textit{Margin Sampling} performs best. In (c), we investigate the sensitivity of the \methodName{} framework to annotator errors by simulating a pixel classification user error (SUE) rate of 10\%. We observe that performance is only marginally affected, indicating the practical robustness of the \methodName{} framework.}
\end{figure*}

\noindent{\textbf{Annotation diversity.}}
Given a fixed pixel labelling budget, 
a natural question arises: \textit{is it better to label a small number of images densely or a large number of images sparsely}? To address this question we design a simple experiment, where a fixed annotation budget of $n$ pixels is
to be distributed over a dataset of $N_\text{total}$ images.
We define the \textit{annotation diversity ratio}, $\eta = \frac{N_{\text{img}}}{N_{\text{total}}}$, 
where $N_{\text{img}}$ refers to the number of images that have had at least one pixel labelled
(for simplicity, we assume the labelling budget is evenly distributed over the selected set of images).
Therefore, $\eta \rightarrow 1$ refers to a budget uniformly distributed over the full dataset (thereby forming a sparse, but diverse, label set),
$\eta \rightarrow 0$ denotes the case where the budget is only spent on a few images (yielding a densely annotated subset of images).
We then train the DeepLabv3+ models on the \cv{} and \cs{}, fixing $B$ so as to end up with 10 pixel labels per image when $\eta = 1$, and experiment with 5 different diversity ratios $\eta$ from 0.01 to 1.0. In Fig.~\ref{fig:exp:data_diveristy},
we observe that mean IoU increases monotonically with $\eta$. 
This indicates that, 
given a fixed budget, it is better to sparsely annotate as many images as possible,
rather than a smaller number more densely, motivating our sparse \methodName{} approach.
In the remaining experiments, we likewise spend our annotation budget evenly across all images within a dataset (as described in Sec.~\ref{subsec:framework}), with each image being only sparsely labelled. \\[-10pt]
 
\noindent \textbf{Encoder depth.}
We next investigate the effect of encoder capacity in the low annotation regime. 
Specifically, we experiment with a ResNet-based FPN by changing the number of layers in the encoder from 18 to 101 layers. 
All encoders are initialised with a model pretrained for classification on ImageNet~\cite{Russakovsky2015ImageNetLS}.  
We conduct experiments both on \cv{} (training each model with 1 to 100 randomly labelled pixel coordinates per image) and \pascalShort{} (training each model with 1 to 1000 randomly sampled labelled pixel coordinates per image), reporting results in~Fig.~\ref{fig:exp:depth_camvid} and Fig.~\ref{fig:exp:depth_pascal}, respectively.
We observe that deeper networks yield higher performance above a minimum number of labelled pixels (approximately 10) per image.
This implies that, at the cost of greater computational complexity, the use of a deeper network may be a viable way to reduce annotation requirements in low annotation regimes (above some minimum labelling threshold).\\[-9pt]

\noindent \textbf{Encoder initialisation.} Next, we investigate whether supervised pretraining is necessary for good segmentation performance in a low annotation regime. Concretely, we compare the performance of an FPN-based architecture with a ResNet50 encoder that is initialised using either supervised (ImageNet classification) or self-supervised~(MoCov2~\cite{chen2020improved}) pretraining. To study how performance differs with the number of labelled pixels, 
we vary the annotation budget from 1 to $10^4$ randomly sampled labelled pixels per image on \cv{}~(Fig.~\ref{fig:exp:self_sup_camvid}) 
and \pascalShort{}~(Fig.~\ref{fig:exp:self_sup_pascal}). 
On \cv, we observe an interesting biphasic phenomenon: when the number of labelled pixels per image is fewer than 10, the model initialised with supervised ResNet50 shows a superior performance. 
However, as the number of pixel labels increases, self-supervised pretraining gradually outperforms its supervised counterpart. 
This phenomenon is also observed in the \pascalShort{} dataset, with a cross-over occurring at approximately $10^2$ labelled pixels per image.
Thus, supervised pretraining may be an appropriate choice for low annotation budgets, when suitable pretraining annotations are readily available, but its advantage wanes the annotation budget grows. Given its superiority at low annotation levels, we adopt supervised pretraining for the remaining experiments.

\begin{figure*}[!htb]
  \centering
    \hspace{-10pt}
  \subfigure[Comparison to prior work on \cs.]
  { 
      \label{fig:exp:sota_cityscape}
   \includegraphics[width=0.3\textwidth,height=0.25\textwidth]{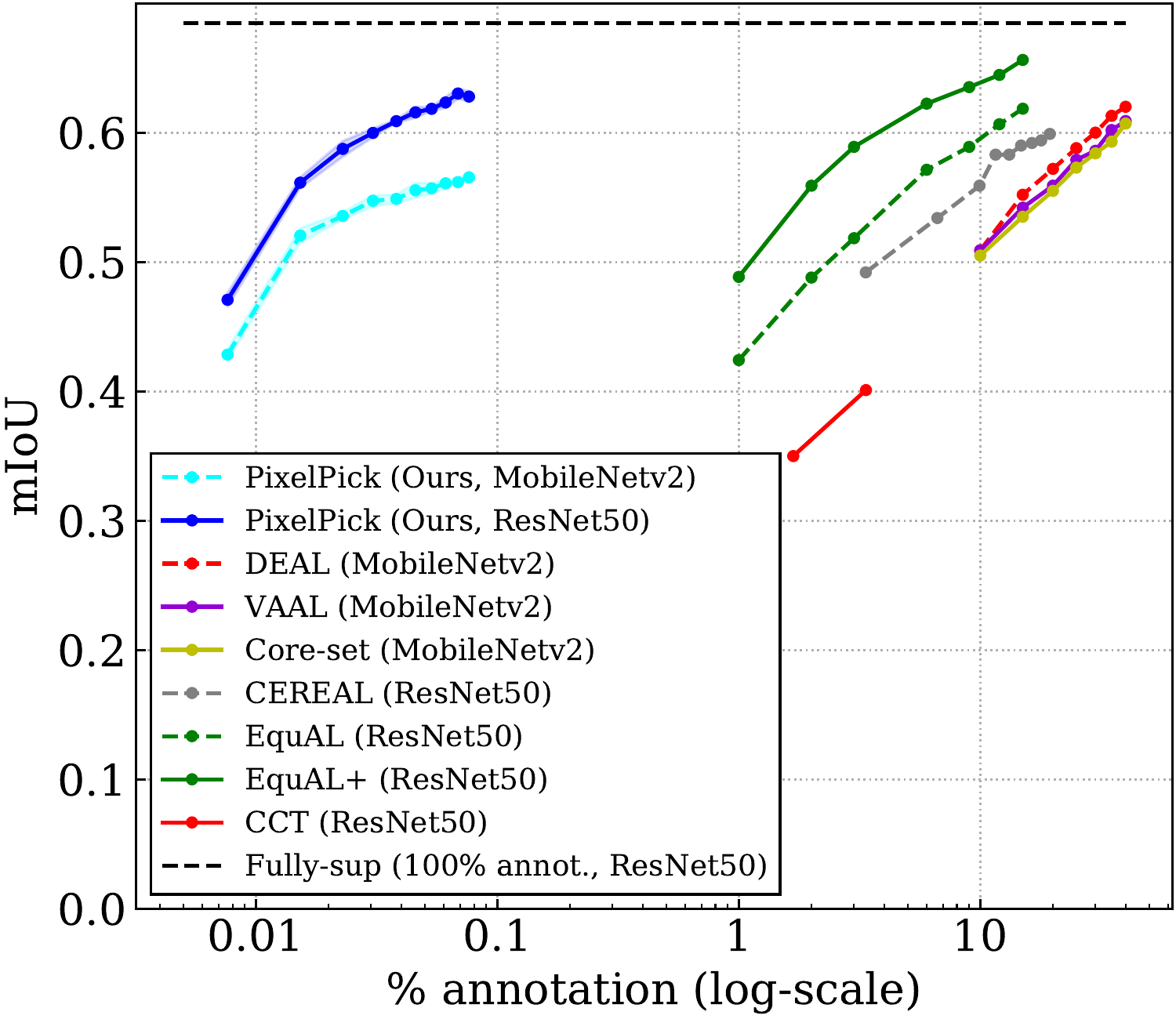}%
  }%
    \hspace{0pt}
  \subfigure[Qualitative results for models trained with \methodName{} on \pascalShort{} (top) and \cs{} (bottom).]
  { 
  \raisebox{0.35cm}{\includegraphics[width=0.695\textwidth,clip,trim={2cm 0cm 0.25cm 0.3cm}]{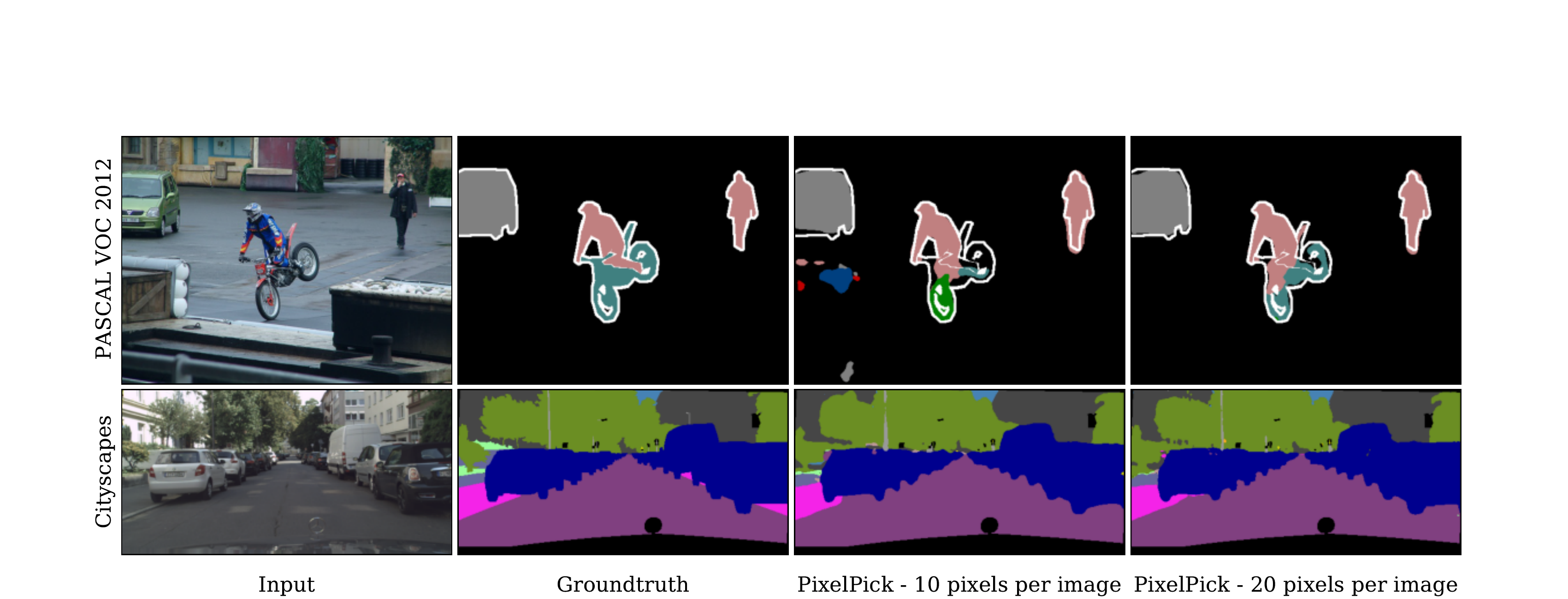}}
  }

 \hspace{-10pt}
  \caption{\textbf{Comparison to state-of-the-art and qualitative results}. In (a) we observe that \methodName{} performs favourably against existing state-of-the-art approaches for active learning and semi-supervised learning on \cs. In (b) we show qualitative results. With only 10 labelled pixels per image, segmentation models trained with \methodName{} achieve promising visual quality, which further improves to capture fine details (e.g. the cleanly segmented thin lamppost in the bottom right image) as further labelled pixels are used.}
  \vspace{-.5cm}
\end{figure*}

\noindent \textbf{Acquisition function.} 
Thus far, we have only labelled pixels selected via simple uniform random sampling,  showing that modern CNNs---with their strong inductive biases---can be trained for semantic segmentation with just a handful of pixel annotations per image. 
Here, we go one step further, investigating whether a better choice of acquisition function can further improve learning efficiency.
To this end,  we experiment on \cv{} with three popular uncertainty sampling methods~(described in Sec.~\ref{subsec:framework}):
\textit{Least Confidence} (LC), \textit{Margin Sampling} (MS) and \textit{Entropy Sampling} (ENT).
In addition, we also experiment with a Query-By-Committee (QBC)~\cite{Seung1992QueryBC} approach 
that queries labels using model ensembles~\cite{settles2009active}. 
We implement this with dropout after each convolutional layer,
repeating inference 20 times to obtain a Monte Carlo estimate following~\cite{Gal16}. 
Due to the large number of models to be trained~(i.e.~different acquisition functions, each trained five times to estimate variance), 
we employ the lightweight MobileNetv2-based DeepLabv3+ model.
We initialise training with 10 uniform randomly selected labelled pixels per image.
Once training converges, we query 10 additional pixel labels with the given acquisition function. 
As described in Sec.~\ref{subsec:framework}, 
we first take top $M$\% ranked pixels (here, $M=5$) per image under the uncertainty estimation ranking  and uniformly sample 10 pixels from these pixels. 
Fig.~\ref{fig:exp:uncertainty} shows the results. 
We see that all uncertainty-based methods outperform the random baseline in every round. 
Interestingly, dropout-based voting variants of LC, MS and ENT each show worse performance than their counterparts voting---a similar observation was also made in~\cite{casanova2020reinforced}.
We note that in our problem setting, \textit{Margin Sampling} (MS) outperforms other strategies, 
reaching about 96\% of the performance of the fully supervised baseline with only 0.06\% of the annotations.  Therefore, we use MS as our sampling method for \methodName{} to compare against previous work in the following section. \\[-10pt]

\noindent \textbf{Discussion.} To summarise, we can draw the following conclusions from the ablation studies:
First, given a fixed pixel annotation budget, it is best to spread it over as many images as possible;
Second, the inductive bias in modern ConvNets makes them well-suited to capture local correlations within an image, 
evidenced by the first three experiments, where models trained with randomly sampled pixels still perform well;
Third, although it might be thought that deeper networks with greater capacity would suffer significantly from over-fitting in the low-annotation regime,  
we found that for many budget choices, deeper networks are the preferred option.
Fourth, in terms of acquisition functions, active learning outperforms random sampling, and in particular, \textit{Margin Sampling} performs best in our setting.

\subsection{Comparison to state-of-the-art methods} \label{subsec:exp:sota}
We next validate our framework by comparing against prior work in active/semi-supervised learning on \cv{}~(Fig.~\ref{fig:teaser}),
\cs{}~(Fig.~\ref{fig:exp:sota_cityscape}) and \pascalLong{}~(Tab.~\ref{tab:exp:sota_voc}). 
To strike a balance between computation complexity and performance, 
we adopt the FPN model with a ResNet50 backbone, 
and query additional samples each round with \textit{Margin Sampling}, 
as suggested by the ablation study. We train for 10 query rounds, with each round adding 10 labelled pixels per image for \cv{} and \cs{} and 5 pixels for \pascalShort{}. Notice that these numbers are far smaller (three orders of magnitude) than the number of pixels required to annotate a single 128x128 size patch as considered in \cite{casanova2020reinforced, mackowiak2018cereals}, 
and not requiring mouse operations, making our approach more efficient.
In each case, 
we observe that \methodName{} is able to achieve comparable performance to the prior state-of-the-art with far fewer pixel annotations (for instance two orders of magnitude on \cv).
We refer the readers to supplementary material for more details on the compared methods.

\begin{table}[!htb]
\small
\setlength{\tabcolsep}{2.pt}
    \centering
    \begin{tabular}{ll|c|c|c|c}\toprule
        Method & Backbone & Fine & Weak & Spatial & mIoU (\%)\\
        & & annot. & annot. & coord. & \\\midrule
        WSSL~\cite{Papandreou_2015_ICCV} & VGG16 & 1.5K & 9K & \cmark & 64.6 \\
        GAIN~\cite{Li_2018_CVPR} & VGG16 & 1.5K & 9K & \cmark & 60.5\\
        MDC~\cite{Wei_2018_CVPR} & VGG16 & 1.5K & 9K & \cmark & 65.7\\
        DSRG~\cite{Huang_2018_CVPR} & VGG16 & 1.5K & 9K & \cmark & 64.3\\
        FickleNet~\cite{Lee_2019_CVPR} & VGG16 & 1.5K & 9K & \cmark & 65.8\\
        BoxSup~\cite{dai2015boxsup} & VGG16 & 1.5K & 9K & \cmark & 63.5\\
        CCT~\cite{Ouali_2020_CVPR} & ResNet50 & 1.5K & 9K & \cmark &\textbf{69.4}\\\midrule
        GAIN~\cite{Li_2018_CVPR} & VGG16 & - & 10.5K & \xmark & 55.3\\
        MDC~\cite{Wei_2018_CVPR} & VGG16 & - & 10.5K & \xmark & 60.4\\
        DSRG~\cite{Huang_2018_CVPR} & ResNet101 & - & 10.5K & \xmark & 61.4\\
        FickleNet~\cite{Lee_2019_CVPR} & ResNet101 & - & 10.5K & \xmark & 64.9\\

        BoxSup~\cite{dai2015boxsup} & VGG16 & - & 10.5K & \cmark & 62.0\\
        ScribbleSup~\cite{lin2016scribblesup} & VGG16 & - & 10.5K & \cmark & 63.1\\\midrule
        PixelPick (Ours) & MobileNetv2 & - & \underline{1.5K} & \cmark & 57.2 \\
        PixelPick (Ours) & ResNet50 & - & \underline{1.5K} & \cmark & \textbf{68.0} \\\bottomrule
    \end{tabular}
    \caption{\textbf{Comparison to existing weakly- and semi-supervised methods on \pascalShort{} validation set.} 
    The third and fourth columns denote the number of fine (dense) and weakly annotated images used for training. The fifth column denotes whether the annotations incorporate a spatial component (for either fine or weak annotation).
    }
    \label{tab:exp:sota_voc}
    \vspace{-.2cm}
\end{table}

\subsection{Practical deployment} 
\label{subsec:exp:practical}
Thus far, we have largely followed the common practice in previous active learning segmentation work, 
mimicking the labelling process by simply disclosing the corresponding labels from the fully-annotated dataset. 
In this section, we evaluate the efficiency of our \methodName{}~(Fig.~\ref{fig:mf-tool}) and its sensitivity to annotator noise during model training.

In detail, 
we ask one annotator to label 100 images from \pascalShort{} dataset, 
with 10 pixels per image,
we measure the average time and accuracy~(between annotator and the groundtruth from original dataset).
As a result, 
with our annotation tool~(despite not being fully optimised), 
it takes less than 1 second on average to label the queried pixel~(10s per image),
with 90\% average accuracy.
To our knowledge, 
this annotation speed is significantly faster than drawing bounding boxes or scribbles~\cite{dai2015boxsup, lin2016scribblesup}, and approximately twice as fast as picking extreme points according to times reported by~\cite{papadopoulos2017extreme}.
Additionally, given the observed annotation error rate, 
we conduct an experiment to assess the influence of these noisy annotations, 
that is, we artificially jitter 10\% of the annotations to simulate errors during the annotation process and train a model on pixels containing this label noise.
As shown in Fig.~\ref{fig:exp:errsim}, 
the performance gap incurred from annotation noise is negligible, indicating that our framework is not only efficient w.r.t.~annotation time but also robust to potential errors caused by annotators.

\section{Conclusion} \label{sec:conclusion}
In this work we proposed \methodName{}, 
a framework for semantic segmentation that employs a small number of sparsely annotated pixels to train effective segmentation models.
We showed that \methodName{} requires considerably fewer annotations than existing state-of-the-art to achieve comparable performance. Finally, we showed how annotation for pixel-level active learning can be obtained efficiently with a mouse-free labelling tool, facilitating real-world deployment.

\vspace{5pt}
\noindent \textbf{Acknowledgements.}  
GS is supported by AI Factory, Inc. in Korea. 
WX and SA are supported by Visual AI (EP/T028572/1). 
The authors would like to thank Tom Gunter for suggestions.
SA would also like to thank Z. Novak and S. Carlson for support.

{\small
\bibliographystyle{ieee_fullname}
\bibliography{refs}
}
\clearpage
\begin{appendices}

\section{Overview} 
In this supplementary material, we present three additional studies:
(i) an evaluation into the effect of varying the number of pixel coordinates sampled in each round of training (Sec.~\ref{sec:effectiveB});
(ii) the influence of our proposed diversity heuristic (Sec.~\ref{sec:diversity}), and
(iii) the effectiveness of a human at selecting pixel coordinates in comparison to using model uncertainty (Sec.~\ref{sec:humanLabelling}).
Finally, we present additional details about methods we compared to in the main paper, that were omitted due to space constraints (Sec.~\ref{sec:otherMethods}).

\section{Effect of the number of queried pixel coordinates per round} \label{sec:effectiveB}
\begin{figure}[!htb]
\centering
  { 
    \hspace{-15pt}
    \includegraphics[width=0.2291\textwidth]{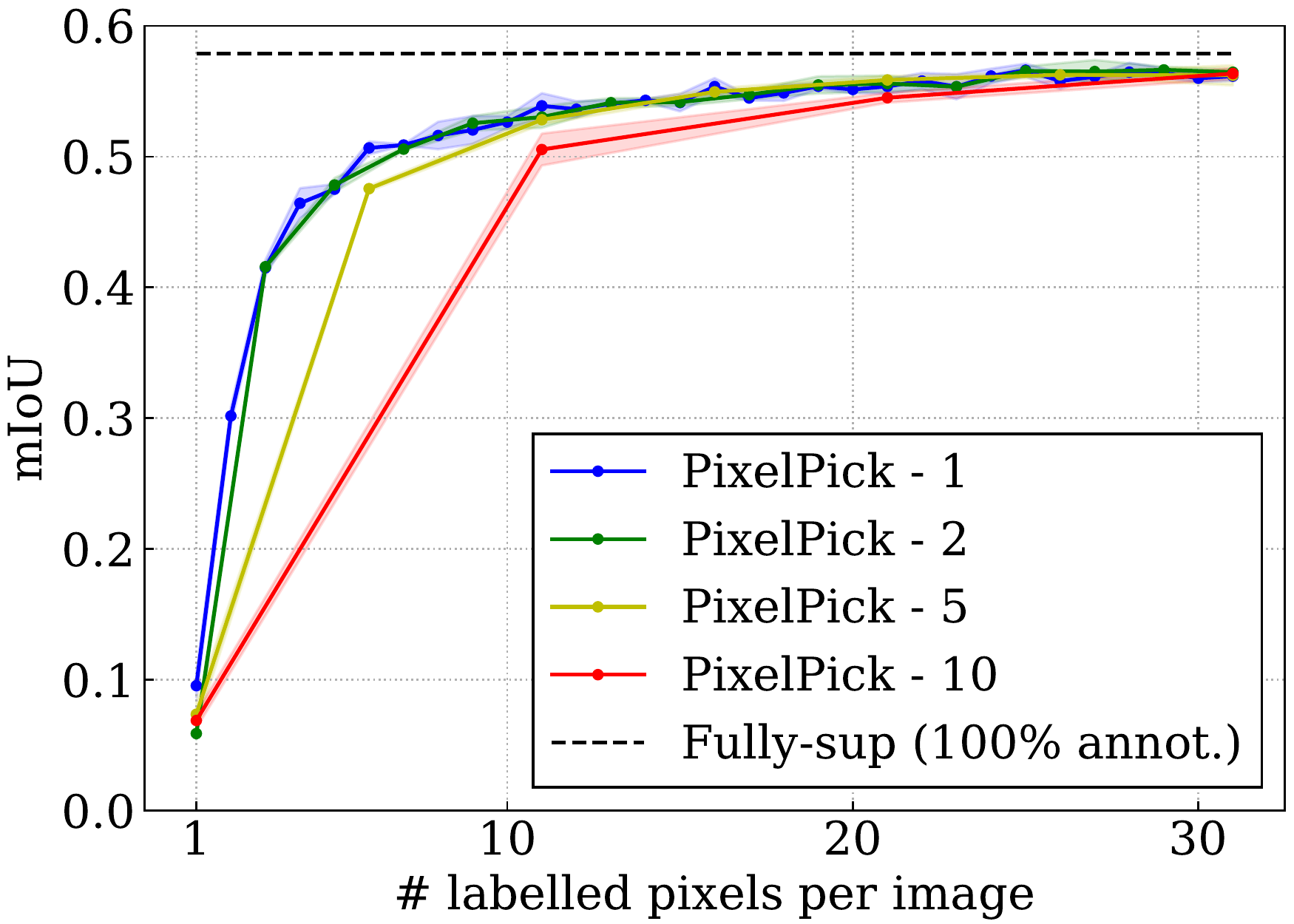}%
  }%
    \hspace{0pt}
  { 
    \includegraphics[width=0.2291\textwidth]{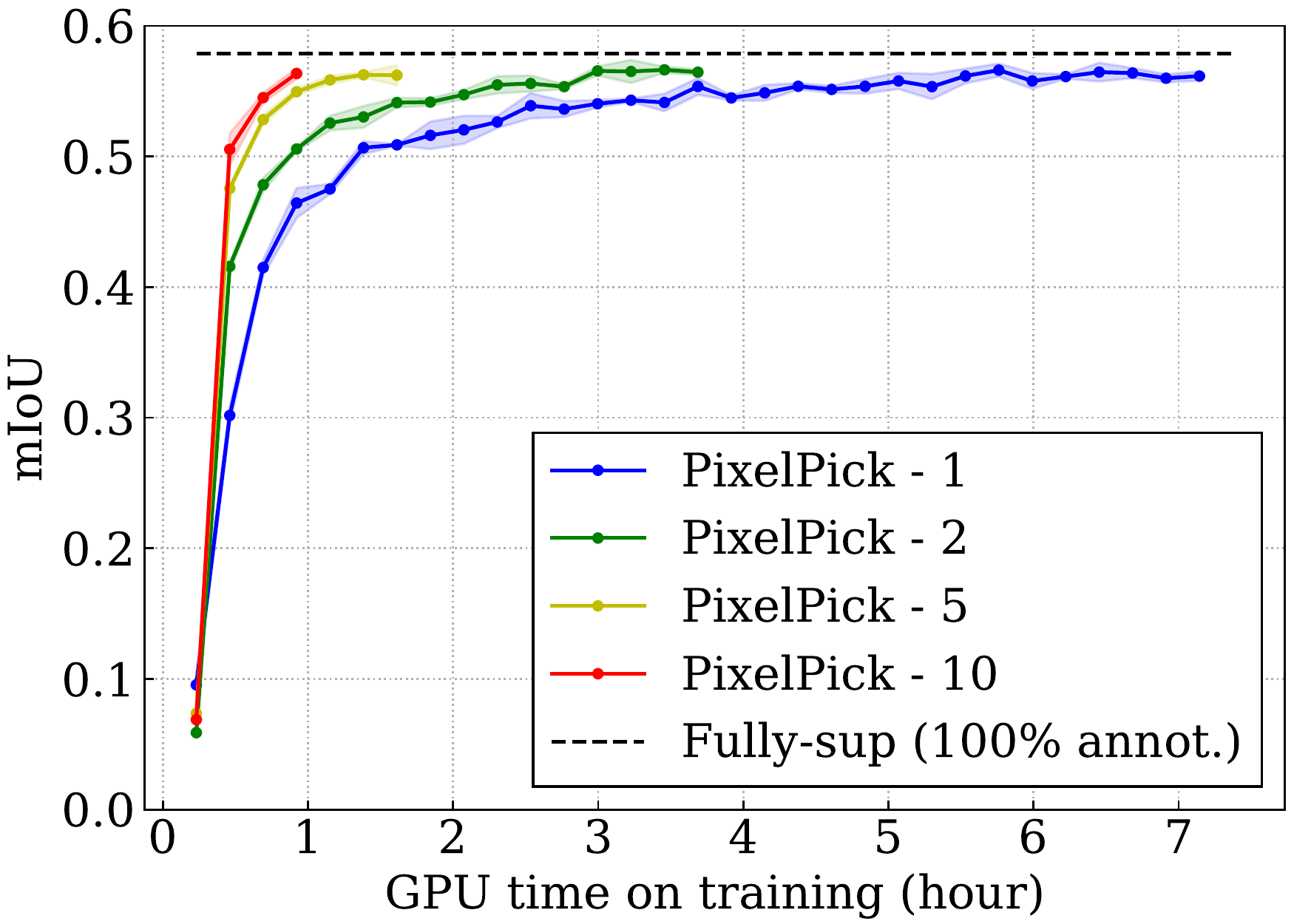}%
  }
 \hspace{-10pt}
 \caption{\textbf{Effect of the number of queried pixel coordinates per round on \pascalShort.} \methodName{}-$n$ denotes our model which samples $n$ pixels per image per query round. Left: given a highly limited annotation budget (e.g., $\leq$ 10 pixels per image), 
 we observe that it is beneficial to pick fewer pixels at each round to achieve a better label efficiency in terms of performance. 
 Right: we show a trade-off between the number of queried pixels per round and total GPU training time taken to reach a certain level of performance.}
 \label{fig:supp_mat:n_pixels_per_round_voc}
 \vspace{-.5cm}
\end{figure}

To understand how the number of labelled pixels added at each round affects the model's performance, we train MobileNetv2-based DeepLabv3+ models on \pascalLong{}. 
Each model queries $n\in$\{1, 2, 5, 10\} pixel(s) per image per round and the maximum budget is set to 30 pixels per image (in the notation employed in Sec.~3 of the paper, $n = B/N$ with $N = 1464$ for the \pascalLong{} dataset). 
All models are given random 1 pixel per image at the beginning of training. 
As shown in Fig.~\ref{fig:supp_mat:n_pixels_per_round_voc} (left), 
we note that when the annotation budget is very low (e.g., $\leq$ 10 pixels per image), 
a model with a lower $n$ value shows a higher mIoU. 
However, when more annotations are allowed (e.g. $\geq$ 20 pixels per image), 
performance is similar across the models.

On the other hand, as the number of query rounds required to reach the max budget is inversely proportional to $n$, we also measure the GPU time for the models to complete the whole training process~(Fig.~\ref{fig:supp_mat:n_pixels_per_round_voc}, right).\footnote{We measure timings on a NVIDIA RTX2080ti GPU card.} We observe that, there is a trade-off between training time and $n$. 
For instance, to reach about 0.5 mIoU, the model has to be re-trained 6 times~(corresponding to an annotation budget of 6 pixels per image) when $n=1$, whereas one would only need to query once~(corresponding to an  annotation budget of 11 pixels per image), if $n = 10$, reducing the overall training time by a factor of 5.

\section{Diversity heuristic} \label{sec:diversity}

As noted in \cite{Beluch2018ThePO,Yoo2019LearningLF}, 
simply selecting samples with the highest uncertainty can result in poor performance due to a lack of diversity among samples. 
In our \methodName{} framework, 
this manifests as querying pixels from a
limited set of spatial regions, 
which is likely to incur redundant queries, 
and in turn degrades the labelling efficiency. 

\begin{figure}[!htb]
\centering
  { 
    \hspace{-15pt}
    \includegraphics[width=0.2291\textwidth]{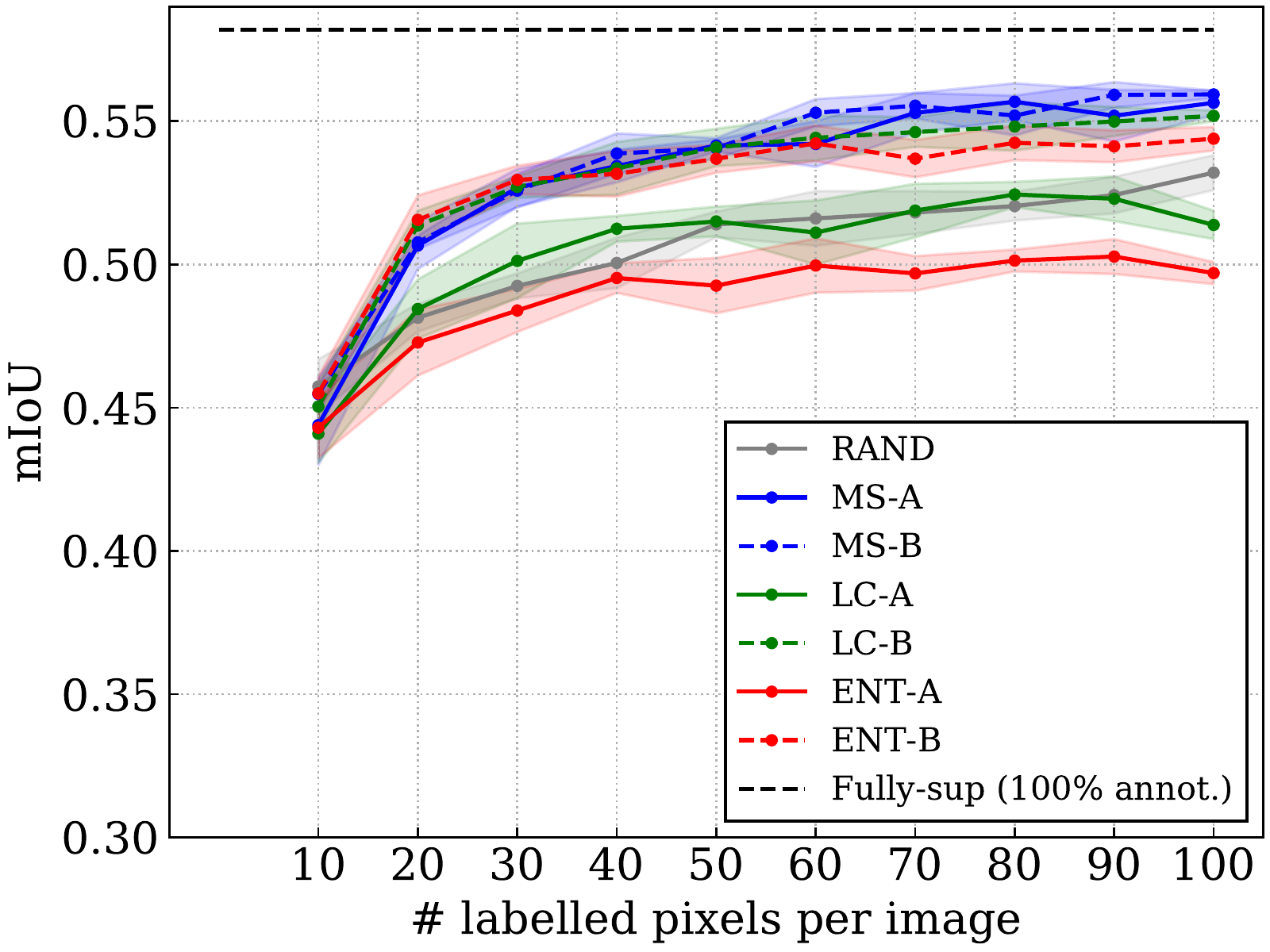}%
  }%
    \hspace{0pt}
  {
    \includegraphics[width=0.2291\textwidth]{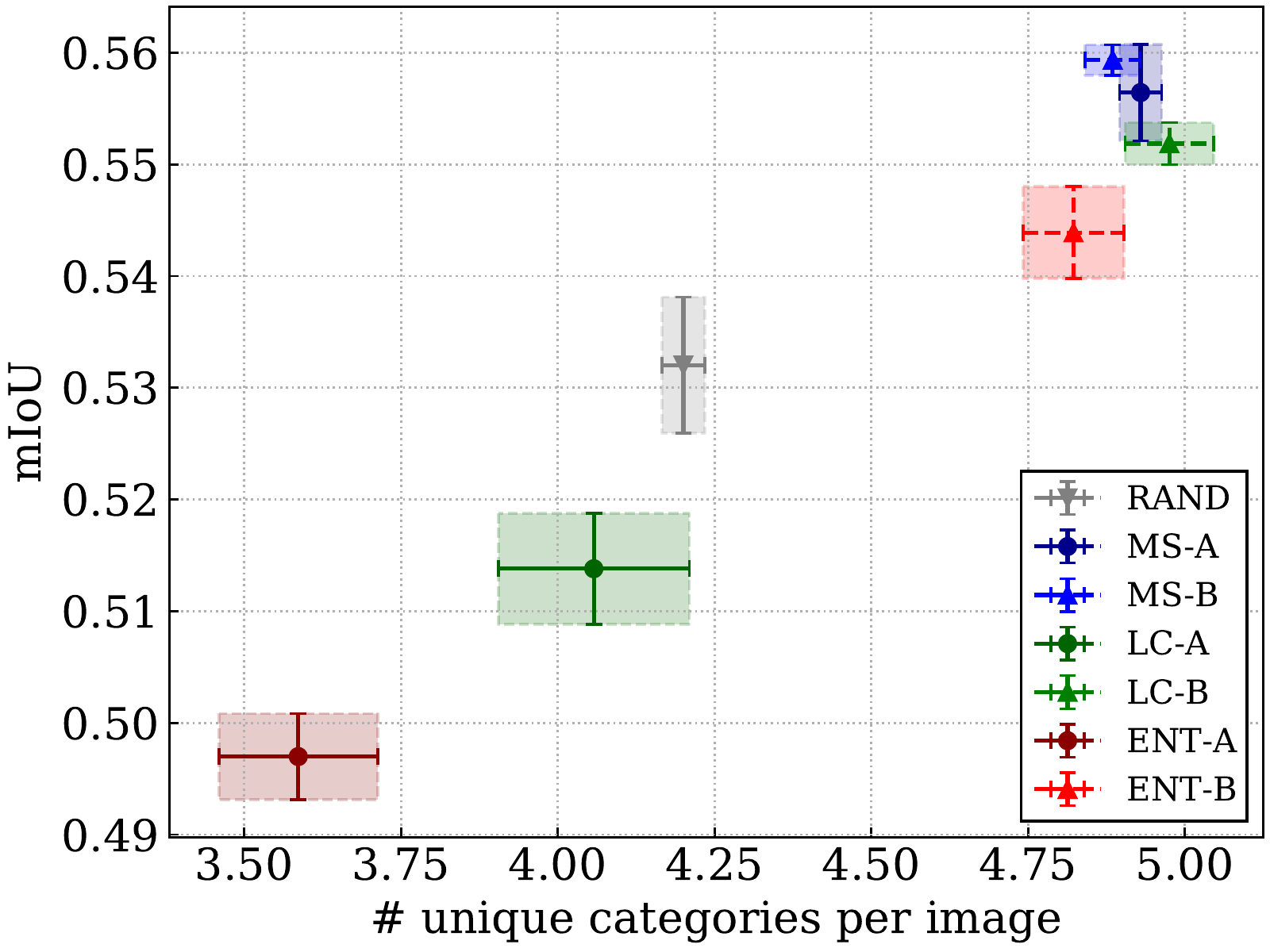}%
  }
 \hspace{-10pt}
 
 \caption{\textbf{Effect of diversity heuristic on \cv{}.} Left: we observe that directly selecting $n$-most uncertain pixels from randomly sub-sampled regions as in~\cite{Yoo2019LearningLF} within an image is sensitive to the choice of an acquisition function (denoted as \{MS, LC, ENT\}-A). In contrast, uniformly choosing $n$ pixels per image from $M$\% pixels with highest uncertainty is robust to the acquisition functions and shows better performance (denoted as \{MS, LC, ENT\}-B). Right: we show that the average class diversity per image covered by the queried pixel locations plays an important role in performance.}
 \label{fig:supp_mat:div_heu}
\vspace{-.5cm}
\end{figure}

To alleviate this effect, 
\cite{Yoo2019LearningLF} sub-sampled the unlabelled pool and chose the $n$-most uncertain samples from the resulting subset. 
We experiment with this approach by uniformly sampling 5\% pixel coordinates within an image and then taking as queries the 10 most uncertain pixels amongst them at each query stage. 
Specifically, we train DeepLabv3+ models on \cv{} for 10 rounds, 
with 10 random labelled pixels per image given at the beginning of training. 
However, as shown in Fig.~\ref{fig:supp_mat:div_heu} (left, denoted by \{MS, LC, ENT\}-A), 
this heuristic does not show promising results compared to the random baseline (RAND) and the performance varies significantly depending on the sampling strategies. 
For example, choosing entropy (ENT-A) as the acquisition function yields a lower mIoU than RAND, whereas using margin sampling (MS-A) allows a better performance. 
We conjecture that this is because directly selecting $n$-most uncertain pixels from the uniformly sub-sampled unlabelled pixels still tends to collect from a few restricted regions (i.e. less diversity). 

Instead, to gather queried pixels from more diverse objects, we propose in the paper to first sample 5\% unlabelled pixels with highest uncertainty and uniformly select 10 pixels from the this subset (denoted as \{MS, LC, ENT\}-B in Fig.~\ref{fig:supp_mat:div_heu}). Put differently, we swap the order of the uniform and uncertainty sampling processes.
As can be seen in Fig.~\ref{fig:supp_mat:div_heu}, the proposed approach brings better results and is robust to the choice of a uncertainty strategy in the pixel-level active learning setting.

To provide evidence for our hypothesis on diversity of the queried pixels, 
we compute the average number of unique categories for queried pixels within an image as an approximate diversity measure. 
As can be seen in Fig.~\ref{fig:supp_mat:div_heu} (right), ENT-A and LC-A, 
which show worse performance than the uniform sampling (RAND) at the end of AL,
queried pixels from less diverse classes than RAND. 
On the other hand, methods with a higher mIoU queried from objects with greater category diversity than RAND, underpinning our hypothesis. We therefore use the proposed diversity heuristic throughout our experiments in the main paper.

\section{Human labelling oracle} \label{sec:humanLabelling}
To show it is beneficial to query labels from the model's perspective rather than a human annotator, 
we compare models trained with labelled pixels selected by one of the uncertainty sampling strategies and by a human annotator. 
For this, we train a MobileNetv2-based DeepLabv3+ on \cv{}, 
given 10 labelled pixels per image queried based on a sampling method and 10 random pixels per image initially offered at the beginning of AL (i.e. retrain after one query round). 
For human-picking, we ask one annotator to pick 10 pixels per image on \cv{} from the regions where the model makes wrong predictions, assuming humans can well recognise the groundtruth annotation from an image, and thus are able to easily validate the errors from the model prediction. The annotator was encouraged to pick pixel coordinates that they believe most useful for boosting segmentation performance from the annotator's view.

Interestingly, as shown in Tab.~\ref{tab:hlc}, we found the performance of the model trained on human-picked pixels is worse than any other uncertain-based strategies, 
even lower than the random baseline by 1.6 mIoU (\%). We found this result surprising---our hypothesis is that human annotators tend to treat each image independently, and consequently tend not to take account of the differing degrees of visual variety present in each class (for example, ``sky'' pixels often look similar, but the ``building'' class can vary significantly in appearance and therefore requires more labels) whereas the model can determine this information readily (via its uncertainty) across the full training set.  The result highlights a potential discrepancy between what really helps the model and what human annotators think useful for solving the task.
A better understanding of the nuances underpinning this effect would be useful future work.

\begin{table}
\begin{center}
\begin{tabularx}{0.3\textwidth}{c|c}
    \toprule
    Sampling method & mean IoU (\%)\\
    \midrule
    Random & 48.1 $\pm~0.5$\\
    Entropy & 51.6 $\pm~0.9$\\
    Least Confidence & 51.4 $\pm~0.5$\\
    Margin Sampling & 50.8 $\pm~0.2$\\
    \midrule
    Human annotator & 46.5 $\pm~0.4$\\
    \bottomrule
\end{tabularx}
\end{center}
\caption{\textbf{Performance comparison between human-picked and uncertainty-based pixels on \cv{}.}}
\label{tab:hlc}
\end{table}

\section{Methods description} \label{sec:otherMethods}
To help readers understand the difference in the methods used for the comparison on \pascalLong{} validation set in our paper, we categorise them according to annotation level they use (i.e., image-, box-, or scribble-level) and briefly summarise each method. We also describe CCT~\cite{Ouali_2020_CVPR}, which primarily addresses semi-supervised learning. All weakly-supervised methods train on \pascalShort{} augmented by SBD~\cite{Bharath2011ICCV} (10.5K images). When they consider semi-supervised setting jointly with their weakly-supervised approach, they use the original \pascalShort{} 1.5K pixel-level annotations for full-supervision and the remaining 9K images for weak-supervision. By contrast, our \methodName{} framework leverages sparse weak-supervision on the 1.5K \pascalShort{} images.

\begin{itemize}
    \item \textbf{Image-level annotation}
    \begin{itemize}
        \item \textbf{WSSL}~\cite{Papandreou_2015_ICCV} adopts an EM-approach in which they estimate segmentation masks given observed image values and image-level labels in the E-step and optimise model parameters on the estimated segmentation in the M-step.
        
        \item \textbf{GAIN}~\cite{Li_2018_CVPR} proposes to use attention maps to enable a better quality of localisation maps for training a segmentation model. To this end, they train an image classification model with an additional attention mining loss to enforce the model to guide itself where to look. To validate their approach, they evaluate another weakly supervised segmentation model, SEC~\cite{Kolesnikov2016ECCV} trained on pseudo-segmentation masks generated by hard-thresholding their attention maps.
        
        \item \textbf{MDC}~\cite{Wei_2018_CVPR} leverages image-level labels to produce pseudo segmentation masks. In particular, they propose to use a convolutional block with multiple dilated rates in order to transfer the discriminative object region to other parts of the object.
        \item \textbf{DSRG}~\cite{Huang_2018_CVPR} uses image-level labels and a deep network pretrained on image classification to produce seed cues which a segmentation network is trained on. The seed cues are further extended to unlabelled pixels by the proposed region growing algorithm in an iterative manner.
        
        \item \textbf{FickleNet}~\cite{Lee_2019_CVPR} generates localisation maps with a pretrained image classification network by saliency, which are further used as pseudo-labels to train a segmentation network. For this, they aggregate a variety of localisation maps, which of each is produced from a single image by applying stochastic hidden unit selection and Grad-CAM~\cite{Selvaraju_2017_ICCV} and highlights different parts of objects present in the image. 
    \end{itemize}
    
    \item \textbf{Box-level annotation}
    \begin{itemize}
        \item \textbf{BoxSup}~\cite{dai2015boxsup} exploits bounding box annotations, 
        which are much easier to obtain than dense pixelwise annotations, at a cost of offering weaker supervision. 
        For this, they iteratively generate semantic masks by forming candidate segments with a unsupervised region proposal method and assigning a semantic label of a groundtruth box to the most overlapped segment and train deep networks on the estimated semantic masks.
    \end{itemize}
    
    \item \textbf{Scribble-level annotation}
    \begin{itemize}
    \item \textbf{ScribbleSup}~\cite{lin2016scribblesup} proposes to use scribble annotations and iterate over propagating them to unmarked regions by optimising a graphical model and training a segmentation model on the generated masks.
    \end{itemize}
    
    \item \textbf{Semi-supervised approach}
    \begin{itemize}
        \item \textbf{CCT}~\cite{Ouali_2020_CVPR} utilises cross-consistency loss to take advantage of unlabelled data under the cluster assumption. For this, they enforce invariance between outputs of auxiliary decoders and main decoder, where the former takes a perturbed embedding from the encoder, and the latter receives clean features from the encoder.
        They train on \pascalShort{} for the fully-supervised pixel-wise cross-entropy loss and on the images from \cite{Bharath2011ICCV} for the cross-consistency loss.
    \end{itemize}
\end{itemize}

\end{appendices}

\end{document}